 \documentclass[authoryear,preprint,review,12pt]{elsarticle}

\usepackage{epsfig}

\usepackage{setspace}
\usepackage{amssymb}
\usepackage{mathtools}
\usepackage{soul}
\usepackage{tabulary, booktabs}
\usepackage{graphicx}

\usepackage{tikz}
\usetikzlibrary{decorations.pathreplacing,angles,quotes,patterns, arrows, arrows.meta,
                chains,patterns.meta,
                positioning,
                shapes.geometric, calc, shapes.multipart}

\usepackage{tikz}
\usetikzlibrary{fit} % Load the fit library for enclosing nodes
\usetikzlibrary{decorations.pathreplacing,angles,quotes,patterns, arrows, arrows.meta,
                chains,patterns.meta,
                positioning,
                shapes.geometric, calc, shapes.multipart}
\usepackage{xcolor}
\usetikzlibrary{arrows}
\usepackage[margin=1in]{geometry} 
\usepackage{xcolor}
\usepackage[utf8]{inputenc}
\usepackage[T1]{fontenc}

\usepackage{bm}     %boldfont
\renewcommand{\vec}[1]{\bm{#1}} %indicate vector by bold font, not by arrow

\usepackage{graphicx, subfig}

\usepackage{subfig}

\usepackage{amsmath,amsthm,amssymb, verbatim, amsfonts}
\usepackage[mathscr]{euscript}
\usepackage{mathtools}
\usepackage{multirow}
\usepackage{color} 
\usepackage{enumitem}       %personalised enumerations
\usepackage{tabulary, booktabs}
\usepackage{caption}
\DeclareMathOperator*{\argmin}{arg\,min}
\usepackage{url}

\usepackage[breaklinks]{hyperref}

\usepackage{subcaption}
\usepackage{tikz,graphics,color,fullpage,float,epsf,caption,subcaption}
\renewcommand{\citet}{\cite}

\graphicspath{ {./_images/} }

\journal{Transportation Research Part E: Logistics and Transportation Review}

\begin{document}

\begin{frontmatter}

\title{Differentiated Pickup Point Offering\\for Emission Reduction in Last-Mile Delivery}

\author[inst1]{Albina Galiullina}
    \ead{albina.galiullina@proton.me, +31 6 3935 3233}
\author[inst2]{Wouter J.A. van Heeswijk}
\author[inst1]{Tom Van Woensel}
\affiliation[inst1]{organization={Department of Industrial Engineering and Innovation Sciences,\\ Eindhoven University of Technology},
            addressline={PO Box 513}, 
            postcode={5600 MB},
            city={Eindhoven},
            country={Netherlands}}
\affiliation[inst2]{organization={Department of Industrial Engineering \& Management Science, Faculty of Behavioural, Management and Social Sciences,\\ University of Twente},
            addressline={Ravelijn  3351}, 
            postcode={7522 NH},
            city={Enschede},
            country={Netherlands}}

% Abstract
\begin{abstract}
Pickup points are widely recognized as a sustainable alternative to home delivery, as consolidating orders at pickup locations can shorten delivery routes and improve first-attempt success rates. However, these benefits may be negated when customers drive to pick up their orders. This study proposes a Differentiated Pickup Point Offering (DPO) policy that aims to jointly reduce emissions from delivery truck routes and customer travel. Under DPO, each arriving customer is offered a single recommended pickup point, rather than an unrestricted choice among all locations, while retaining the option of home delivery. We study this problem in a dynamic and stochastic setting, where the pickup point offered to each customer depends on previously realized customer locations and delivery choices. To design effective DPO policies, we adopt a reinforcement learning–based approach that accounts for spatial relationships between customers and pickup points and their implications for future route consolidation. Computational experiments show that differentiated pickup point offerings can substantially reduce total carbon emissions. The proposed policies reduce total emissions by up to 9\% relative to home-only delivery and by 2\% on average compared with alternative policies, including unrestricted pickup point choice and nearest pickup point assignment. Differentiated offerings are particularly effective in dense urban settings with many pickup points and short inter-location distances. Moreover, explicitly accounting for the dynamic nature of customer arrivals and choices is especially important when customers are less inclined to choose pickup point delivery over home delivery.
\end{abstract}
\end{frontmatter}

% % \begin{graphicalabstract}
% % \includegraphics{compensationCost_on_assignment_distance}
% % \end{graphicalabstract}

% % \begin{highlights}
% % \item Research highlight 1
% % \item Research highlight 2
% % \end{highlights}
% %% keywords here, in the form: keyword \sep keyword

% %% PACS codes here, in the form: \PACS code \sep code

%%%%%%%%%%%%%%%%%%%%%%%%%%%%%%%%%%%%%%%%%%%%%%%%%%%%%%%%%%%%%%%%%%%%%%
%                         INTRODUCTION
%%%%%%%%%%%%%%%%%%%%%%%%%%%%%%%%%%%%%%%%%%%%%%%%%%%%%%%%%%%%%%%%%%%%%%

\section{Introduction}\label{section:introduction}
As e-commerce grows, last-mile delivery becomes more challenging, accounting for 53\% of total shipping costs \citep{accenture_faster_last_mile} and 13\% of total city carbon emissions \citep{weforum}. To improve the efficiency and sustainability of the last mile, service providers adopt alternative delivery methods to complement traditional attended home delivery. One promising alternative is pickup point delivery, in which orders are delivered to intermediate locations, such as stores, post offices, or automated lockers, for customer collection.

Pickup points offer advantages for both service providers and customers. For providers, pickup points enhance last-mile efficiency by consolidating orders into a few pickup points, reducing the number of stops required for delivery and lowering the risk of failed deliveries \citep{morganti2014impact, deutsch2018parcel,janjevic2019integrating, rautela2022investigating}. Customers may also find this option more convenient than waiting at home for deliveries, which can be restrictive and uncertain \citep{civitas}. This growing preference for pickup point delivery is evident in recent trends: the share of home delivery in Europe declined to 54\% in 2025 \citep{dhl_outofhome}, while the number of pickup point locations increased by 22\% between 2023 and 2025 \citep{gel_proximity}.

Despite these advantages, the overall environmental impact of pickup point delivery remains ambiguous when customers’ travel emissions are considered. For example, \citet{niemeijer2023greener} show that when a pickup point is located more than 3 km from the customer's home, the probability that the customer makes a round trip by car exceeds 30\%. Consequently, because of such customers' travel, pickup point delivery may result in higher total carbon emissions than home delivery, contrary to the intended sustainability benefits of pickup point delivery. These findings highlight that the sustainability of pickup point delivery depends critically on how pickup points are offered to customers, rather than on their mere availability.

Taking into account both the delivery truck and customers' travel emissions, this study examines differentiated pickup point offerings as a tool to minimize total last-mile carbon dioxide emissions. We consider a next-day delivery service within a defined region, with multiple pickup points and a single depot served by a single truck.  Orders arrive throughout the day, and customers choose between home delivery and pickup at a designated point when placing orders. Unlike industry practice, where customers are given unrestricted choice among all pickup points, we examine the policy of offering a single recommended pickup point per order. This policy reflects a realistic operational intervention, as service providers increasingly influence customer choices through recommendations, default options, and interface design. Given the pickup offer, the customer decides whether to accept pickup point delivery or opt for home delivery. The next day, the truck fulfills the delivery route based on customers' choices. Next, customers who select pickup point delivery travel to collect their orders using either non-polluting modes (e.g., walking or cycling) or car travel.  The objective is to design a differentiated pickup point offering (\textit{DPO}) policy that dynamically selects one pickup point for each arriving customer, to minimize total carbon emissions from both delivery operations and customer travel.

Determining which pickup point to offer each customer is not trivial because it requires balancing the trade-off between order consolidation and customer proximity to pickup points. Offering the nearest pickup point or home delivery minimizes customer travel-related emissions but can increase delivery truck emissions by requiring trucks to visit multiple locations. Conversely, consolidating orders at a few centrally located pickup points reduces truck emissions but increases customer travel distances and associated emissions.  The problem is inherently dynamic, because each decision influences not only the current order but also the structure of future deliveries. For example, all else equal, offering a pickup point that is already planned to be visited is more efficient than introducing a new stop.

The dynamic trade-off between consolidation and proximity gives rise to a sequential decision-making problem under uncertainty, which can be naturally modeled as a Markov decision process (MDP). When a customer arrives, the policy assigns a pickup point based on the current system state, while accounting for its impact on future consolidation. Although the action space is small (selecting one pickup point from a fixed set), the state space is large and variable, as it captures geographical relations between a dynamically changing set of customers and pickup points. To capture these geographical and relational dependencies in a scalable manner, we represent the system state as a graph with tailored connections and node features. This representation allows the policy to account for spatial proximity and consolidation structure while accommodating a variable number of customers. We employ a graph attention mechanism to prioritize relevant local interactions, such as nearby customers or already-visited pickup points, while retaining global information needed to estimate future routing costs. Policy learning is performed using Proximal Policy Optimization (PPO), a well-established reinforcement learning algorithm.

Our numerical analysis demonstrates the effectiveness of the proposed differentiated pickup point offering policy across a range of operational settings. The resulting policy consistently outperformed a PPO baseline with a flattened state representation and a graph-based model without attention. We further conduct experiments on problem settings to identify when a DPO policy is most beneficial and when simpler alternative policies perform adequately. In particular, we examine how the policies adapt to changes in the size of the service region, customer choice probabilities, and the time of order arrival. 

Our contributions can be summarized as follows:
\begin{enumerate}
\item We introduce a new dynamic policy for differentiated pickup point offerings that aims to minimize total last-mile carbon emissions. The policy explicitly accounts for emissions from both delivery trucks and customer travel, and captures the trade-off between consolidation efficiency and customer proximity to the pickup points.
\item We propose a scalable graph-based state representation that captures the spatial and consolidation structure of the system and enables effective learning of dynamic pickup point offering policies under uncertainty.
\item We provide a systematic numerical study that identifies problem settings in which differentiated offerings are most effective, and those in which simpler policies, such as home-only delivery, unrestricted pickup point choice, or nearest pickup point assignment, perform reasonably well. Differentiated offerings are most valuable in dense urban environments with many pickup points and short inter-location distances, and when customers are less inclined to choose pickup point delivery.
\item We benchmark the proposed policy against a perfect-information solution and develop a dynamic heuristic that captures key insights from the learned policy while reducing computational complexity.
\end{enumerate}

The remainder of the paper is organized as follows. Section~\ref{section:related_literature} reviews the relevant literature.
Section~\ref{section:dps} describes the problem and Section~\ref{section:formulation} describes formulation. The solution methodology is presented in Section~\ref{section:methodology}. Section~\ref{section:instance_design_setup} describes the setup for the numerical experiments, and Section~\ref{section:computational_experiments} analyzes the numerical experiments, and the decisions of the learned policy.  Section~\ref{section:conclusion} concludes the paper.

\section{Related literature}\label{section:related_literature}
This section reviews two research streams relevant to our study. Section~\ref{section:lit_rev_impact_pups} reviews the impact of pickup points on the sustainability of last-mile delivery. Section~\ref{section:lit_rev_dm_in_lm} reviews customer choice modeling and control in last-mile delivery.

\subsection{Impact of pickup points on the sustainability of last-mile delivery \label{section:lit_rev_impact_pups}}

Pickup points are recognized as an effective approach for solving last-mile delivery challenges. We refer to reviews by \citet{rohmer2020guide}, \citet{janinhoff2024out}, and \citet{mohri2024contextualizing} that explore various aspects of pickup point delivery. Our focus is specifically on the sustainability benefits of pickup point delivery, such as reducing the failed delivery rate, decreasing delivery truck travel distances, and enhancing the robustness of delivery systems.

Failed home deliveries remain a major operational concern, particularly when customers are unavailable to receive parcels. Pickup points mitigate this issue by serving as primary delivery locations or as backup options when initial home delivery fails. Empirical studies in the UK \citep{mcleod2009quantifying, song2009addressing, edwards2010carbon, mcleod2006transport, song2013quantifying} indicate that pickup points can significantly reduce delivery truck routes by minimizing redelivery attempts. They also reduce customers' travel distances by eliminating the need to retrieve packages from distant depots when redelivery is unsuccessful.

Pickup points enable the consolidation of orders at fewer locations, reducing the distance delivery trucks travel. \citet{vogt2007tabu} formalize this as a routing-allocation problem, in which customers are divided into two groups: those receiving home deliveries and those assigned to visited vertices (e.g., pickup points), to minimize the combined routing and allocation costs. They propose a tabu search heuristic for solving this problem. In a multi-depot, two-echelon variant of this problem, \citet{Zhou2018} use a hybrid genetic algorithm to optimize routes and allocations.  This problem can also yield emission reductions by treating routing costs as truck emissions and allocation costs as customers' travel emissions. Focusing on minimizing total delivery emissions, \citet{goodchild2018analytical} use a continuous approximation approach to estimate last-mile emissions, comparing two extreme cases: when all orders are delivered to homes versus when all are directed to pickup points. Using real-world data from a logistics provider in Poland, \citet{iwan2016analysis} compare these scenarios, reporting a tenfold reduction in truck emissions when all orders are directed to pickup points rather than homes.

Multiple studies have focused on optimizing pickup point networks, examining their locations and capacities \citep{deutsch2018parcel, janjevic2019integrating, lin2020last, xu2021data, rautela2022investigating, peppel2022impact, peppel2024integrating, zadeh2026strategic}. \citet{wang2025does} further emphasize that network density and infrastructure are critical to making pickup point delivery operationally efficient.  A case study by \citet{van2020home} further indicates that pickup points can improve robustness to operational uncertainty while reducing truck emissions.
 
Although customers often show a willingness to use pickup points for sustainability reasons \citep{caspersen2021sharing, caspersen2022act, biancolin2024environmental}, customers' delivery preferences and transportation choices significantly affect environmental benefits. Studies across various geographical locations indicate that pickup points tend to improve last-mile delivery sustainability in urban areas, whereas in rural and suburban regions, home delivery is more sustainable due to limited access to pickup points. For example, case studies in Belgium \citep{rai2020consumers, beckers2021sustainability, mommens2021delivery} demonstrate that more than half of the population cannot reach pickup points within walking distance.  Moreover, even in capital city regions with a high density of pickup points, nearly half of customers use cars to collect their orders, as evidenced by studies in Australia \citep{collins2015behavioural}, Sweden \citep{liu2019assessing}, and Belgium \citep{rai2020consumers}. Using real shipment data from Oslo, \citet{pinchasik2023lockers} show that although replacing home delivery with parcel-locker delivery can reduce total carbon dioxide emissions by about 13--32\%, the net benefits are highly sensitive to customers’ pickup-trip mode and travel distance.  \citet{niemeijer2023greener} conclude in their study conducted in the Netherlands that customers' travel emissions can be so significant that it may be more sustainable to limit pickup point delivery to customers living within walking distance or even to remove this option in regions where delivery trucks have transitioned to renewable fuel.

To summarize, the literature shows that pickup point delivery can substantially reduce last-mile emissions through route consolidation. Still, these gains depend on customer locations, travel behavior, and the design of the pickup point network. Existing studies largely rely on real-world case analyses, whereas formal optimization models typically assume static settings, such as unrestricted access to all pickup points or comparisons between full home delivery and full pickup point delivery. What remains unexplored is a dynamic, operational perspective that accounts for the sequential arrival of customers and stochastic customer choices. This gap motivates our study, which proposes a differentiated pickup point offering policy that explicitly captures these dynamic trade-offs.

\subsection{Customer choice control in last-mile delivery \label{section:lit_rev_dm_in_lm}}
Steering customer choices to improve operational efficiency is commonly achieved through demand management tools. It has been shown to enhance both environmental performance through reductions in emissions, waste, and energy consumption \citep{agatz2023demand}, as well as to increase revenues and reduce operational costs (see \citet{klein2020review}, \citet{wassmuth2023demand}, and \citet{fleckenstein2023recent} for recent reviews). In the context of last-mile delivery, two main approaches have received significant attention: time slot management and delivery option offering.

In time slot management, service providers influence customer demand across delivery windows to balance workloads over time and consolidate geographically proximate orders that might otherwise be assigned to separate routes. Demand steering is typically implemented through differentiated delivery fees, restricted time slot availability, or monetary incentives for rescheduling deliveries \citep{Campbell2006IncentiveServices, agatz2011time, Yang2016, klein2019differentiated, estrada2019biased, ulmer2020dynamic, yildiz2020pricing}. Closely related are approaches that differentiate shipping fees based on customers’ geographic locations, thereby shaping demand geographically to maximize service profit \citep{afsar2021vehicle, afsar2022traveling}.

A second line of work focuses on the delivery option offering, in which service providers steer demand toward pickup point delivery rather than home delivery. \citet{Zhou2018}, \citet{tilk2021last}, and \citet{dumez2021large} study this problem in static, deterministic settings, in which the provider jointly selects the delivery mode (home delivery or pickup point delivery) and the delivery time slot for each order while accounting for customer preferences. These studies employ a hybrid genetic algorithm \citep{Zhou2018}, an exact branch-price-and-cut algorithm \citep{tilk2021last}, and a large neighborhood search heuristic \citep{dumez2021large}, respectively. Extending this line of research, \citet{galiullina2024demand} consider a static stochastic setting in which customers are incentivized in advance to choose pickup point delivery, although their final choices remain uncertain. They propose exact and heuristic branch-and-bound methods to solve this problem.

Only a limited number of studies address customer choice control in dynamic and stochastic environments. \citet{akkerman2024learning} model delivery option selection as a Markov decision process and use deep reinforcement learning to learn a pricing policy that influences customer choice between home delivery and one of the five nearest pickup points. Their approach yields an 8.1\% cost reduction compared to a static discounting policy. \citet{mancini2025dynamic} study the assignment of pickup points to customers to maximize provider profit. Different from our work, they assume mandatory acceptance of assigned pickup points and allow penalized delivery delays, without accounting for distance-dependent customer acceptance or environmental impacts.

The delivery option offering has also been implemented to manage limited pickup point capacity. \citet{wang2025prompting} propose penalizing late pickups to improve locker availability. \citet{sailer2024dynamic} study a dynamic and stochastic setting in which pickup points are selectively excluded from the choice set to account for uncertain pickup times and locker occupancy. In contrast, our approach restricts the offer to a single pickup point per customer and explicitly models customer acceptance as a function of distance, thereby capturing the trade-off between consolidation benefits and customer travel emissions in a dynamic setting.

To summarize, we contribute to the literature in three ways. First, we study pickup point offering in dynamic settings with sequential customer arrivals and stochastic acceptance behavior, which remain largely unexplored in the literature. Second, we explicitly account for environmental trade-offs by incorporating distance-dependent customer travel emissions alongside delivery routing emissions. Third, from a methodological perspective, we propose an algorithmic framework for the dynamic–stochastic pickup point offering problem that captures uncertainty, delayed cost realization, and interdependent customer decisions. 

\section{Differentiated pickup point offering for next-day delivery}\label{section:dps}

We consider a next-day delivery service within a defined region, with multiple pickup points and a single depot served by a truck.  The delivery planning process consists of two phases: 
\begin{enumerate}
    \item \textbf{Capture phase} occurs during the day before delivery until the order acceptance cut-off time. During this phase, orders arrive dynamically. For each new order, a single pickup point is offered, and customers randomly choose between the pickup point and home delivery. The number and the location of upcoming orders are uncertain.
    \item \textbf{Fulfillment phase} occurs on the delivery day when the delivery truck's route is executed. All order and customer choice information is determined.
\end{enumerate}

During the capture phase, upon receiving a customer's order, the provider determines which delivery options to offer. Home delivery is always offered, and the provider may also offer a single pickup point location as an alternative. The customer then randomly selects between home and pickup point delivery, with their choice influenced by the proximity of the offered pickup point. Customers closer to the pickup point are more likely to choose that option. The provider uses information from existing customers during the current capture phase to inform pickup point offering decisions for new orders.

The fulfillment phase starts at the order acceptance cut-off time. First, a route is constructed based on each customer's chosen delivery option, either to the customer's home or the selected pickup point. After the truck delivers the order, customers who have opted for pickup point delivery travel to the pickup point to collect their orders. Customer transportation modes can be categorized as non-polluting (e.g., walking, cycling) or polluting (e.g., car trips) with respect to carbon dioxide emissions. Additionally, car trips that are done for other purposes than solely order collection, referred to as ``chained" trips, are considered non-polluting. The likelihood of choosing a polluting transportation mode is positively correlated with the distance between the customer's home and the offered pickup point; greater distances increase the probability of choosing polluting modes.

During the fulfillment phase, there is an interaction between two primary sources of carbon dioxide emissions in last-mile delivery: \textit{(i)} the delivery truck and \textit{(ii)} customers' travel to pickup points. Emissions from the delivery truck are directly proportional to the length of its route. In contrast, customers' travel emissions depend on the probability of each customer choosing a car trip for order collection, which is nonlinearly influenced by the distance to the offered pickup point. This study aims to develop a policy offering a single pickup point to each customer, to minimize the total carbon dioxide emissions of both emission sources.

\subsubsection*{Example of the delivery planning process\label{sec:example}}
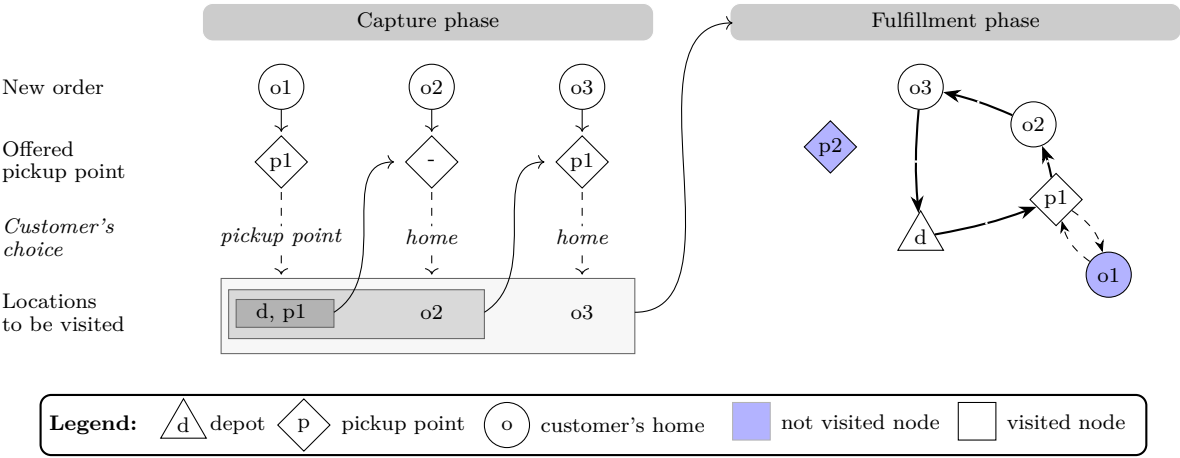
\begin{figure}[h!]
\begin{tikzpicture}[font=\scriptsize, depot/.style={regular polygon, regular polygon sides=3, minimum size=0.7cm, inner sep=0pt, draw=black},
customer_visit/.style={circle,  minimum size=0.6cm, inner sep=0pt, draw=black}, 
pup_visit/.style={diamond, minimum size=0.7cm, inner sep=0pt, draw=black},  
customer_notvisit/.style={circle, fill=blue!30, minimum size=0.6cm, inner sep=0pt, draw=black}, 
pup_notvisit/.style={diamond, minimum size=0.7cm, inner sep=0pt,  fill=blue!30,  inner sep=0pt, draw=black}]
    \node at (-1,3) [label=right:{New order}](requests) {};
    \node at (-1, 2) [label=right:{\shortstack[l]{Offered\\pickup point}}] {};
    \node at (-1,1) [label=right:{\shortstack[l]{\textit{Customer's}\\\textit{choice}}}]  {};
    \node at (-1,0) [label=right:{\shortstack[l]{Locations\\to be visited}}]  {};
%%%%%%%%%%%%%%%%%%%% titles of phases   %%%%%%%%%%%%%%%%%%%%%
\node (aux_title) [above = 0.57cm of requests] {};
\node (aux_title2) [right  = 0.01cm of aux_title] {};
 \node [right = 2.5cm of aux_title2,draw=white, rectangle,inner sep=0.05cm, rounded
corners,fill=black!20,minimum width=6cm, minimum height=0.5cm] (title_capture_phase) {Capture phase};
 \node [right = 1cm of title_capture_phase,draw=white, rectangle,inner sep=0.05cm, rounded
corners,fill=black!20,minimum width=6cm, minimum height=0.5cm] (title_fulfillment_phase) {Fulfillment phase};

%%%%%%%%%%%%%%%%%%%%%%%% requests %%%%%%%%%%%%%%%%%%%%%%%%
\node[customer_visit] at ((3, 3) {o1};
\node[customer_visit] at ((5, 3) {o2};
\node[customer_visit] at ((7, 3) {o3};
    \foreach \x in {3, 5, 7} {
        \draw[->] (\x, 2.7) -- (\x, 2.4);
    }
%%%%%%%%%%%%%%%%%%%%%%%% retilaer's decisions %%%%%%%%%%%%5
     \node[diamond, draw, minimum size=0.7cm] at (3, 2) [label=center:{p1}] {};
    \node[diamond, draw, minimum size=0.7cm] (rd_o2) at (5, 2) [label=center:{-}] {};
    \node[diamond, draw, minimum size=0.7cm] (rd_o3)  at (7, 2) [label=center:{p1}] {};
    \foreach \x in {3, 5, 7} {
        \draw[->, dashed] (\x, 1.6) -- (\x, 0.5);
    }
    \node[draw=white, rectangle, fill=white] at (3, 1) [label=center:{\textit{pickup point}}] {};
     \node[draw=white, rectangle, fill=white] at (5, 1) [label=center:{\textit{home}}] {};
      \node[draw=white, rectangle, fill=white] at (7, 1) [label=center:{\textit{home}}] {};
% Rectangle around Customer's decisions
\draw[rectangle,  draw=black!60, fill=black!3] (2.2, -0.55) rectangle (7.7, 0.45) ;
\draw[rectangle, draw=black!60, fill=black!15] (2.3, -0.35) rectangle (5.7, 0.3);
\draw[rectangle, draw=black!60, fill=black!30] (2.4, -0.2) rectangle (3.7, 0.17) ;

%%%%%%%%%%%%%%%%%%%%%%%% customer's decisions %%%%%%%%%%%%5
    \node at (3, 0) [label=center:{d, p1}] {};
    \node at (5, 0) [label=center:{o2}] {};
    \node at (7, 0) [label=center:{o3}] {};

% Arrow from rectangle to route
\draw[->]  (7.7, 0)  .. controls +(1.5,0) and +(-1.5,0).. (9, 3.85);
\draw[->] (5.7, 0)  .. controls +(0.8,0.5) and +(-0.8, 0).. (6.5, 2);
\draw[->] (3.7, 0)  .. controls +(0.8,0.5) and +(-0.8, 0)..  (4.5, 2);

%%%%%%%%%%%%%%%%%%%%%%%%  fulfillment phase %%%%%%%%%%%5
 \begin{scope}
   \node[depot] (d1) at (11.5, 1) {d};
    
    \node[customer_visit] (o1) at (11.5, 3) {o3};
    \node[customer_visit] (o2) at (13, 2.5) {o2};
    \node[customer_notvisit] (o3) at  (14, 0.5)  {o1};
    
    %\node[pup_notvisit] (p1) at   (14.5, 3) {p2};
    \node[pup_notvisit] (p2) at   (10.3, 2.2) {p2};
    \node[pup_visit] (p1) at  (13.3, 1.5) {p1};

    %%%% route %%%%
 \begin{scope}[>={Stealth[black]},
              every node/.style={fill=white,circle,inner sep=0.1pt},
              every edge/.style={draw, thick, bend right=5}]
     \path [->] (o1)  edge node {} (d1);
     \path [->] (o2)  edge node {} (o1);
     \path [->] (p1)  edge node {} (o2);
      \path [->] (d1)  edge node {} (p1);
\end{scope}

 %%%% customer - pup arrows %%%
  \begin{scope}[>={Stealth[black]},
              every node/.style={fill=white,circle,inner sep=0.1pt},
              every edge/.style={draw,dashed, bend left=20}]
     \path [->] (o3)  edge node {} (p1);
     \path [->] (p1)  edge node {} (o3);
\end{scope}

\end{scope}
\begin{scope}
 \node at (0.5, -1.5) [label=center:{\textbf{Legend:}}] (legend) {};
\node[depot, label=right:{depot}]  (depot) at (1.7, -1.5) {d};
\node[pup_visit, label=right:{pickup point}] (pup_visit) at (3.3, -1.5) {p};
    \node[customer_visit, label=right:{customer's home}] (cust_visit) at (6, -1.5) {o};
    % Enclose all nodes within a rectangle
    \draw[thick, rounded corners] (-0.2, -1.9) rectangle (14.5, -1.1);

\draw[ draw=black!30, fill=blue!30]  (9, -1.7) rectangle (9.5, -1.2); % Draw the rectangle
\node at (9.5, -1.45) [anchor=west] {not visited node}; % Add the label

\draw[draw=black] (12, -1.7) rectangle (12.5, -1.2); % Draw the rectangle
\node at (12.5, -1.45) [anchor=west] {visited node}; % Add the label
\end{scope}
\end{tikzpicture}
\caption{Delivery planning process for three sequential orders}\label{fig:workflow}
\end{figure}

This section demonstrates the delivery planning process using an example with three orders (Figure~\ref{fig:workflow}). During the capture phase, three customer orders arrive sequentially:
\begin{enumerate}[itemsep=0pt]
    \item For the first customer (o1), the provider offers a choice between pickup point (p1) and home delivery. The customer selects the pickup point, requiring both the depot and p1 to be visited during the fulfillment phase.
    \item The second customer (o2) is offered home-only delivery, because a minimal detour is required for this option.
    \item When the third customer (o3) arrives, the provider, already aware of the necessary visits to the depot, p1, and o2, offers a choice between the third pickup point (p3) and home delivery. The customer opts for home delivery.
\end{enumerate} 
 During the fulfillment phase, a delivery truck must visit the depot, the first pickup point (p1), and the homes of the second and third customers (o2 and o3). This differentiated pickup point offer demonstrates route optimization by avoiding remote locations such as (o1) while not offering distant pickup points such as (p2).

\section{Problem formulation}\label{section:formulation}
The problem is defined within a circular service region of radius $L$ containing a set $\mathcal{M}$ of pickup point locations and a depot denoted by vertex $\{0\}$. We define $\mathcal{N}$ as the set of potential customers, which is unknown and is used only for exposition.  The distance between any two locations $i,j \in \{0\}\cup \mathcal{M} \cup \mathcal{N}$ is denoted by $c_{ij}$. The delivery planning process considers orders placed during the capture phase, which begins at time 0 and ends at a cutoff time $T$. Customer orders arrive according to a stochastic process with arrival rate $\lambda$. 

The problem is modeled as a finite-horizon Markov decision process (MDP), detailed in the following subsections. Table \ref{tab:TableOfNotation} describes the main notation used throughout the problem description.

\begin{table}[h!]
\caption{Main notation\label{tab:TableOfNotation}}
  \renewcommand{\arraystretch}{0.8}
\begin{tabular}{lp{0.76\linewidth}}
\toprule
\multicolumn{2}{l}{ \underline{\textit{Global parameters}}}\\
$\mathcal{M}$  & set of pickup points\\
$\mathcal{N}$  & set of potential customers' locations\\
$\{0, \dots, K\}$  & set of decision epochs indexed by $k\in \{0, \dots, K\}$\\
$T$ & cut-off time, defining the order capture phase $[0, T)$\\
$L$ & the radius defining the circular service region\\
$c_{ij}$ & distance between nodes $i$ and $j$, $i,j \in \{0\}\cup \mathcal{M} \cup \mathcal{N}$\\
$e^{\text{car}}$, $e^{\text{truck}}$ & carbon dioxide emission per unit distance associated with customers' cars and delivery trucks, respectively\\ 
$p_{om}$  &probability that customer $o \in \mathcal{N}$ chooses pickup point delivery if  pickup point $m\in\mathcal{M}\cup \emptyset$ is offered\\
$p_{o}(l|m)$  &probability  that customer $o \in \mathcal{N}$ chooses transportation mode $l$, $l \in \{\text{car}, \text{walk}, \text{cycle}, \text{chained}\}$, to collect order from pickup point $m\in\mathcal{M}$\\
\multicolumn{2}{l}{ \underline{\textit{State variables}}}\\
$s_k$  & state of decision epoch $k\in \{0,\dots, K\}$, where $s_k = (\mathcal{V}_k, \mathcal{H}_k, \mathcal{A}_k)$\\
$\mathcal{V}_k$  & set of nodes indexed by $i \in \{0, 1, \dots, |\mathcal{V}_k|\}$\\
$\mathcal{A}_k$  & set of arcs\\
$\mathcal{H}_k$  & set of feature vectors $\vec{h_i} \in \mathcal{H}_k$ of nodes $i\in\mathcal{V}_k$, where $\vec{h_i}=(d_i, t_i, \zeta^{\text{pup}}_i, mv_i)$ \\
$d_i$ & location of the node $i\in \mathcal{V}_k$\\
$t_i$  &time at which a new order arrives, i.e., when the decision must be made (non-zero only for the new order)\\
$\zeta^{\text{pup}}_i$ & binary indicating if the node  $i\in \mathcal{V}_k$ is a pickup point\\
$mv_i$ & binary indicating if the node  $i\in \mathcal{V}_k$ must be visited during the fulfillment  phase\\
\multicolumn{2}{l}{ \underline{\textit{Decision variables}}}\\
%$\mathcal{Z}(s_k)$  &set of decisions in state $s_k$\\
$\vec{z_k} = (z_{km})_{m\in\mathcal{M}}$  &binary vector indicating if pickup point $m\in\mathcal{M}$ is offered to the new customer $o_k$, $k \leq K$; node $m_k^*\in\mathcal{M}\cup\emptyset$ denotes the offered pickup point\\
$\vec{x} = (x_{ij})_{i,j\in\mathcal{V}_K}$  & routing variable,  where $x_{ij} = 1$ if  arc $(i,j), i,j \in \mathcal{V}_K$ is traversed by the truck\\
\multicolumn{2}{l}{ \underline{\textit{Exogenous parameters}}}\\
$o_k$  & information about the new customer, $k\in\{0, \dots, K+1\}$\\
$\omega_{k}(\vec{z_k})$  & delivery option choice of the new customer $o_k$, $k\in\{0, \dots, K\}$\\
\multicolumn{2}{l}{ \underline{\textit{Objective}}}\\
$r_k(s_k, \vec{z_k}, \omega_k(\vec{z_k}))$&cost-to-go for a decision epoch $k\in\{0, \dots, K\}$\\
$ r^{\text{customer}}(\omega_{k}(\vec{z_k}))$ & emission of customer travel to pickup new order $o_k$, $k\in\{0, \dots, K\}$\\
$r^{\text{truck}}(s_K)$  &emission of delivery truck route in final period $K$\\
\bottomrule
\end{tabular}
\end{table}

\subsection{Decision epochs} A decision epoch $k \in \{0,1,\dots, K\}$ is triggered by the arrival of a new order ($k\leq K$). The order capture phase ends when the cutoff time $T$ is reached.  When considering stochastic dynamic order arrival, the number of decision epochs $K$ is also stochastic.   

\subsection{States} 
At each decision epoch $k$, the state $s_k$ is denoted as a graph, where a feature vector represents each node. Specifically, $s_k = (\mathcal{V}_k, \mathcal{H}_k, \mathcal{A}_k)$, where: 
\begin{itemize}
    \item  $\mathcal{V}_k$ is the set of nodes that includes: \textit{(i) }the depot $\{0\}$, \textit{(ii)} the set of pickup points $\mathcal{M}$, \textit{(iii)} the set of customers who placed orders before the decision epoch $k$, $\mathcal{N}_k$, and \textit{(iv)} the new customer's home location $\{o_k\}$. The initial state $s_0$ includes only the vertices of the pickup points and the depot: $\mathcal{V}_0=\mathcal{M}\cup \{0\}$.
    \item $\mathcal{H}_k$ is the set of feature vectors $\vec{h_i}$ for each node $i \in \mathcal{V}_k$. Each vector $\vec{h_i} = (d_i, t_i, \zeta^{\text{pup}}_i, mv_i)$ includes four components: $d_i$ represents the location of the node, $t_i \in [0,T)$ is the arrival time of the new order (set to 0 for other nodes), $\zeta^{\text{pup}}_i$ is $1$ if the node is a pickup point and $0$ otherwise, and $mv_i = 1$ if node $i$ must be visited during the fulfillment phase (i.e., the node is the depot, a pickup point chosen by a customer, or a customer opting for home delivery), and $0$ if node $i$ does not have to be visited.
    \item $\mathcal{A}_k$  is the arc set, which includes: \textit{(i)} self-loops for pickup points and the new order: $ \{(i,i) \mid i \in \mathcal{M}\cup \{o_k\}\}$, \textit{(ii)} directed arcs originating from nodes that must be visited and terminating at pickup points or the new order: $\{(i,j) \mid  i \in \mathcal{V}_k, mv_i = 1, j \in \mathcal{M} \cup \{o_k\}, j \neq i\}$, \textit{(iii)} directed arcs from customers opted for pickup point delivery to the designated pickup point: $\{(i,j)| i \in \mathcal{N}_k, mv_i = 0, j \in \mathcal{M}\}$. % Set  $\mathcal{A}_k$ defines  adjacency matrix $A$, in which  $A_{ij} = 1$ if $(i,j)\in\mathcal{A}$ and $0$ otherwise
\end{itemize}

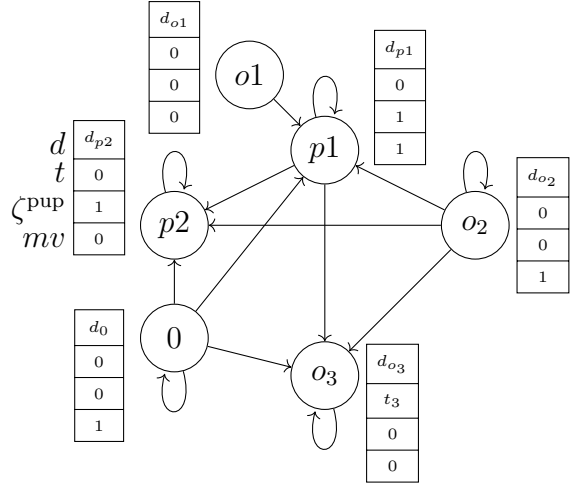
\begin{figure}[h]
\centering
\begin{tikzpicture}[
state_node/.style={circle,  minimum size=0.9cm, inner sep=0pt, draw=black}]
\tikzstyle{bplus}=[rectangle split,rectangle split parts=4, draw, minimum width=0.65cm,
    font=\tiny]
%%%%%%%%%%%%%%%%%%%%%%%%  fulfillment phase %%%%%%%%%%%5
 \begin{scope}

     \node[state_node] (o1) at (12, 2.5) {$o1$};
    \node[state_node] (p1) at (13, 1.5) {$p1$};
    \node[state_node] (p2) at (11, 0.5) {$p2$};
    \node[state_node] (d) at (11, -1) {$0$};
    \node[state_node] (o2) at  (15, 0.5)  {$o_2$};
    \node[state_node] (o3) at (13, -1.5) {$o_3$};

    \node[bplus] (split) at (11, 2.6) {$d_{o1}$\nodepart{two}$0$\nodepart{three}$0$\nodepart{four}$0$};
    \node[bplus] (split) at (14, 2.2) {$d_{p1}$\nodepart{two}$0$\nodepart{three}$1$\nodepart{four}$1$};
    \node[bplus] (split) at (10, 1)  {$d_{p2}$\nodepart{two}$0$\nodepart{three}$1$\nodepart{four}$0$};
     \node[bplus] (split) at (10, -1.5) {$d_{0}$\nodepart{two}$0$\nodepart{three}$0$\nodepart{four}$1$};
      \node[bplus] (split) at  (15.9, 0.5) {$d_{o_2}$\nodepart{two}$0$\nodepart{three}$0$\nodepart{four}$1$};
       \node[bplus] (split) at (13.9, -2)  {$d_{o_3}$\nodepart{two}$t_3$\nodepart{three}$0$\nodepart{four}$0$};
\node[anchor=east] at (9.7, 1.6) {$d$};
\node[anchor=east] at (9.7, 1.2) {$t$};
\node[anchor=east] at (9.7, 0.7) {$\zeta^{\text{pup}}$};
\node[anchor=east] at (9.7, 0.3) {$mv$};
 
    \draw[<->] (p1) edge[loop above] node {} (p1);
    \draw[<->] (p2) edge[loop above] node {} (p2);
    \draw[<->] (o2) edge[loop above] node {} (o2);
    \draw[<->] (o3) edge[loop below] node {} (o3);
    \draw[<->] (d) edge[loop below] node {} (d);

     \draw[->] (o1) -- (p1);
    \draw[->] (o2) -- (p1);
    \draw[->] (o2) -- (p2);
    \draw[->] (o2) -- (o3);
    \draw[->] (p1) -- (p2);
    \draw[->] (p1) -- (o3);
    \draw[->] (d) -- (p1);
    \draw[->] (d) -- (p2);
    \draw[->] (d) -- (o3);

\end{scope}

\end{tikzpicture}
\caption{Example of a state representation with feature vectors as node representations}\label{fig:state_example}
\end{figure}

The state representation for the example presented in Section~\ref{sec:example} is illustrated in Figure~\ref{fig:state_example}, which captures the state at the decision epoch triggered by the arrival of the third customer ($o_3$). There, the must-visit nodes include the depot ($0$), the first pickup point ($p1$), and the second customer ($o2$).

\subsection{Decisions} 
In each epoch $k\leq K$, the provider offers a pickup point location for a new customer. This decision is denoted by a binary vector $\vec{z_{k}} = (z_{km})_{m\in\mathcal{M}}$, where $z_{km} = 1$ if pickup point $m\in\mathcal{M}$ is offered and $0$ otherwise. The offered pickup point is denoted by $m^*_k \in \mathcal{M} \cup \emptyset$, where $\emptyset$ indicates that home-only delivery is offered (that is, when $\vec{z_{k}} = (0)_{m\in\mathcal{M}}$). The decision space $\mathcal{Z}$ does not depend on the state and is defined as:
    \begin{align}
        \mathcal{Z}:=\left\{(z_{km})_{m\in\mathcal{M}}\Big| \sum_{m\in\mathcal{M}}z_{km} \leq 1; \quad z_{km}\in\{0,1\} \;\forall m\in\mathcal{M}, \forall k\in \{0, \dots, K\}\right\}.
    \end{align}

In the final epoch $K$,   triggered when time $T$ is reached, routing decisions are made using an off-the-shelf solver and are not part of the learned policy.  These decisions are represented by a binary vector $\vec{x} = (x_{ij})_{i,j\in\mathcal{V}_K, i\neq j}$, where $x_{ij} = 1$ if the delivery truck traverses arc $(i,j)$, and $0$ otherwise.

\subsection{The exogenous information} 

Exogenous information includes two components: \textit{(i)} the choice of the current customer between the pickup point and home delivery, and \textit{(ii)} the information about the next customer. 

The customer's choice between a pickup point and home delivery is random and influenced by the location of the offered pickup point $m^*_k \in \mathcal{M} \cup \emptyset$, determined by the provider's decision $\vec{z_k}$. Let $p_{o_k m_k^*}$ denote the probability that customer $o_k$ chooses delivery to pickup point $m_k^*\in \mathcal{M}$ over home delivery. For notational purposes, let $p_{o_k m_k^*}\equiv 0$ for  $m^*_k =\emptyset$ denote the probability of choosing the pickup point when home-only delivery is offered. The customer's choice is denoted by a discrete random variable $\omega_k$:
        \[
        \omega_k(\vec{z_k}) = \begin{cases}
        m^*_k, & \text{with probability } p_{o_km^*_k},\\
        o_k, & \text{with probability } 1- p_{o_km^*_k}.
       \end{cases}
        \] 

To model the dependency of probability $p_{o_k m_k^*}$ on the distance to the offered pickup point, we use a multinomial logit framework within a utility-based approach \citep{ben1999discrete, lin2020last}. The utilities of the pickup point and home delivery are defined as follows:
\begin{align}
&u^{\text{pup}}_{m_k^*}= u_0^{\text{pup}} - \beta^{\text{pup}} c_{o_km_k^*} + \epsilon_{o_km_k^*}  &&\text{pickup point delivery,}\\
&u^{\text{home}}= u_0^{\text{home}} + \epsilon_{o_k} &&\text{home delivery,}
\end{align}
where $u_0^{\text{pup}}$ and $u_0^{\text{home}}$ are base utilities, $c_{o_km_k^*}$ is the distance to the pickup point, $\beta^{\text{pup}}$ is a disutility coefficient for customer travel to the pickup point, and $\epsilon_{o_km_k^*}$ and $\epsilon_{o_k}$ are error terms. Parameters $u_0^{\text{pup}}$, $u_0^{\text{home}}$, and $\beta^{\text{pup}}$ can be derived from survey data, as reported in studies such as \citet{haase2009discrete}, \citet{hood2020sociodemographic}, and \citet{milioti2020modelling}. % i.i.d. Standard Gumbel random variables. 
Based on these utilities, the customer's choice probability $p_{o_km_k^*}$ is defined as:
\begin{alignat}{3}   \label{eq:probability_pup}
&p_{o_km_k^*}  = \frac{\exp{(u_0^{\text{pup}}  - \beta^{\text{pup}} c_{o_km_k^*})}}
    {\exp{(u_0^{\text{pup}} - \beta^{\text{pup}} c_{o_km_k^*})} + \exp{(u_0^{\text{home}})}}.
\end{alignat}

The second component of exogenous information is the information about the next customer $o_{k+1}$, represented by a feature vector $h_{o_{k+1}} = (x_{o_{k+1}}, y_{o_{k+1}}, t_{o_{k+1}}, 0, 0)$. If $o_{k+1} = \emptyset$, it signifies the termination of the order capture phase.

%Given the probability \eqref{eq:probability_pup},  $\omega_k$ takes the value $m^*_k$ with probability $p_{km^*_k}$, and the value $o_k$ with probability $1 - p_{km^*_k}$.

\subsection{Transitions}
Unless the order capture phase has terminated ($o_{k+1} = \emptyset$), the state $s_k$ is updated to the next state $s_{k+1}$ in two steps based on exogenous information.  First, a preliminary update accounts for the arrival of the new order $o_{k+1}$ -  the new order node and its features are added to the node set and the set of feature vectors: $\mathcal{V}'_{k} = \mathcal{V}_{k}\cup \{o_{k+1}\}$,  $\mathcal{H}'_{k} = \mathcal{H}_{k} \cup \{h_{o_{k+1}}\}$. The set of arcs is modified by adding a self-loop for $o_{k+1}$, and removing all the arcs connected to the node $o_k$: $\mathcal{A}'_{k} = \mathcal{A}_{k}\cup\{(o_{k+1}, o_{k+1})\} \setminus \{(i, o_{k})| i\in \mathcal{V}_k\}$.  

Second, we update the state based on the current customer's choice realization, $\omega_k$. If the customer opts for pickup point delivery ($\omega_k \in \mathcal{M}$), then the feature vector of the chosen pickup point $m^*_k$ is updated, taking into account that it must be visited ($mv_{m^*_k} = 1$). The arc set is updated by adding arcs originating from $m^*_k$ and terminating at other pickup points and $o_{k+1}$ if those did not exist before. The node $o_k$ is connected to the selected pickup point by a single edge.  If the customer opts for home delivery ($\omega_k = o_k$), the set of nodes and feature vectors remains unchanged. The set of arcs is updated by adding a self-loop for $o_k$, and arcs originating from $o_k$ and terminating at the pickup points and $o_{k+1}$.  These transitions can be formally expressed as:
\begin{equation*}
s_{k+1} = (\mathcal{V}_{k+1}, \mathcal{H}_{k+1}, \mathcal{A}_{k+1}) =
\begin{cases}\!
  \begin{aligned}[c] 
    \Big(
    &\mathcal{V}'_k  \setminus\{o_{k}\}, \quad 
    \mathcal{H}'_{k} \setminus \{h_{k}\}, \\[-1em] 
    &\mathcal{A}'_{k} \cup \{(m^*_k, i) \mid i \in \mathcal{M} \cup \{o_{k+1}\}, (m^*_k, i) \notin \mathcal{A}'_k\} \Big),
\end{aligned}
  &\omega_k(z_k) \in \mathcal{M},\\
\begin{aligned}[c] 
    \Big(
    &\mathcal{V}'_k,\quad 
    \mathcal{H}'_{k}, \\[-1em] 
    &\mathcal{A}'_{k} \cup \{(o_{k}, o_{k})\} \cup \{(o_k, i) \mid i \in \mathcal{M} \cup \{o_{k+1}\}\} \Big),
\end{aligned}
  &\omega_k(z_k) = o_k.
\end{cases}
\end{equation*}

\subsection{The cost-to-go}\label{sec:cost_to_go}
We consider two sources of carbon emission: customers' travel to pickup points and delivery truck routes. Customer travel emissions, indicated by $r^{\text{customer}}(\omega_{k}(\vec{z_k}))$, are calculated at each decision epoch $k \leq K$, while truck emissions, denoted by $r^{\text{truck}}(s_K)$, are computed only at the final epoch $K$. The cost-to-go for each epoch $k \in {0, \dots, K}$, denoted as $r_k(s_k, \vec{z_k}, \omega_k(\vec{z_k}))$, is calculated as $r_k(s_k, \vec{z_k}, \omega_k(\vec{z_k})) = r^{\text{customer}}(\omega_{k}(\vec{z_k}))$ for $k < K$, and $r_K(s_K, \vec{z_K}, \omega_K(\vec{z_K})) = r^{\text{customer}}(\omega_{K}(\vec{z_K})) + r^{\text{truck}}(s_K)$ for the final epoch. Next, we detail the calculation of each cost component.

The emissions from customers' travel, $r^{\text{customer}}(\omega_{k}(\vec{z_k}))$, depend on the realization of customer's choice, $\omega_{k}(\vec{z_k})$, between home and pickup point delivery. If the customer opts for home delivery, travel emissions are zero. However, for pickup point delivery, emissions depend on the transportation mode chosen by the customer: either polluting (e.g., car) or non-polluting (e.g., walking, cycling). Specifically, the emissions for pickup point delivery equal the probability that the customer uses a car for order collection multiplied by the emissions associated with that trip.

Let customer $o_k$ be offered a pickup point $m_k^*$, and let $p_{o_k}(\text{car}|m_k^*)$ denote the probability that the customer chooses a car trip for order collection. The emissions from a car trip are calculated as a round trip between the customer's home and the offered pickup point, given by 
$2 e^{\text{car}} c_{o_k m^*_k}$, where $e^{\text{car}}$ represents the carbon dioxide emission of the customer's car per unit distance. The emissions from the customer's travel are then computed as:
        \[
        r^{\text{customer}}(\omega_{k}(\vec{z_k})) = 
        \begin{cases}
            2 e^{\text{car}} c_{o_k m^*_k} p_{o_k}(\text{car}|m^*_k), & \text{if } \omega_k(\vec{z_k})\in \mathcal{M},\\
            0, & \text{if } \omega_k(\vec{z_k}) = o_k.
        \end{cases}
        \]

The probability $p_{o_k}(\text{car}|m^*_k)$ depends on the location of the offered pickup point.  Following \citet{niemeijer2023greener},  we model this probability as part of the customer’s choice of transportation mode $l$, where $l \in \{\text{car}, \text{walk}, \text{cycle}, \text{chained}\}$. Here, ``chained'' refers to car trips undertaken for other purposes, which are excluded from travel emissions calculations. The utility of using mode $l$ to collect an order is given by $u_{o_k}^l = u_0^l + \beta^l c_{o_km_k^*}  + \epsilon_{o_k}^l$, where $u_0^{l}$ and $\beta^{l}$ are the base utility and the distance sensitivity attributed to mode $l$,  and  $\epsilon_{o_k}^l$ is an error term. The probability of customer $o_k$ choosing  mode $l$ when a pickup point $m_k^*$ is offered is:
\begin{alignat}{3}  \label{eq:prob_car_np} 
&p_{o_k}(l|m_k^*)  &&= \frac{\exp{(u_0^{l} + \beta^{l} c_{o_km_k^*})}}
    {\sum_{l'\in\{\text{car},\; \text{walk},\; \text{cycle},\; \text{chained}\}}\exp{(u_0^{l'} + \beta^{l'} c_{o_km_k^*})}}.
\end{alignat}

The truck route emissions, $r^{\text{truck}}(s_K)$, are calculated at the final epoch $K$ by solving a routing subproblem with an off-the-shelf solver. Thus, routing decisions are implicitly defined by the final state $s_K$. The subproblem minimizes the carbon emission $e^{\text{truck}} c_{ij}$ for each traversed edge $(i,j)$, $i,j\in\mathcal{V}_K, i\neq j$, where $e^{\text{truck}}$ is the carbon dioxide emission of the delivery truck per unit distance. The truck visits only the nodes $i\in \mathcal{V}_K$ that must be visited (indicated by $mv_i = 1$ in $\mathcal{V}_K$), namely, the depot, customer homes for customers opting for home delivery, and chosen pickup points. The subproblem is defined as a traveling salesman problem:
    \begin{alignat}{4}
    & r^{\text{truck}}(s_K) = \quad e^{\text{truck}} && \min && \sum_{i,j \in \mathcal{V}_{K}, i<j}{c_{ij}x_{ij}}&&\label{eq:tsp1}\\
    & && \text{s.t.} && \sum_{\{i,j\} \in \upsilon(\{i\})} x_{ij} = 2	\quad &&\forall i \in \mathcal{V}_{K} \text{ if } mv_{i} = 1\label{eq:eq:tsp2}\\
    & && &&\sum_{\substack{\{i,j\} \in \upsilon(\mathcal{Q}) }} x_{ij}\geq 2 &&\forall \mathcal{Q} \subset \mathcal{V}_{K}, |\mathcal{Q}|\geq 2 \label{eq:eq:tsp3}\\
    & && && x_{ij} \in \{0,1\}	&&\forall i,j \in \mathcal{V}_{K}, i<k\label{eq:eq:tsp4},
    \end{alignat}
where  for some subset of nodes $\mathcal{Q}\subseteq \mathcal{V}_K$, we define $\upsilon(\mathcal{Q})=\{\{i,j\}| i\in \mathcal{Q},j\in \mathcal{V}_K\setminus \mathcal{Q}\}$.

\subsection{The objective and a solution} 
The objective is to minimize the total carbon dioxide emission of both customers' travel and the delivery truck. 

A solution for the problem is given by a decision policy $\pi\in \Pi$ and a decision rule $Z^{\pi}:s_k\rightarrow\mathcal{Z}$ that offers a pickup point or no pickup point option $m_k^* \in \mathcal{M}\cup \emptyset$ for a new customer given current state $s_k$.    The objective is to find an optimal policy $\pi^* \in \Pi$ that minimizes the total expected carbon dioxide emission:
    \begin{align}\label{eq:objective}
        \pi^* = \argmin_{\pi \in \Pi}\mathbb{E}\left[ \sum_{k=0}^{K}r^{\text{customer}}_k( \omega_{k}(\vec{z_k})) + r^{\text{truck}}(s_K)\Big|s_{0},\ \vec{z_k} = Z^{\pi}(s_k)\right].
    \end{align}
    %\begin{align}
    %     V_k(s_k) &= \min_{z}\sum_{m\in \mathcal{M}} p_1(z) \left(\beta_{ie}l_{ie}  + V_{t+1}(x + p) + (1 - p_1(z))V_{k+1}(x+i)\right),\\
    %     &= \min_{z}\sum_{m\in \mathcal{M}} p_1(z) \left(\beta_{ie}l_{ie}  + \left[V_{k+1}(x + p) - V_{k+1}(x+i)\right]\right) +  V_{k+1}(x+i),
    % \end{align} 
    % with boundary condition:
    % \begin{align}
    %     V_{k +1}=  CVRP(\mathcal{V}_k)
    % \end{align}

\section{Solution methodology} \label{section:methodology}

This section describes the architecture of the deep reinforcement learning (DRL) approach used in this work. First, we provide a high-level overview of the components and the information flow between them. Next, Section~\ref{sec:network_architecture} describes the neural network architecture of the agent, focusing on the construction of node embeddings and the Graph Attention Network (GAT) encoder. Section~\ref{sec:ppo} briefly outlines the Proximal Policy Optimization (PPO) algorithm used to train the agent.

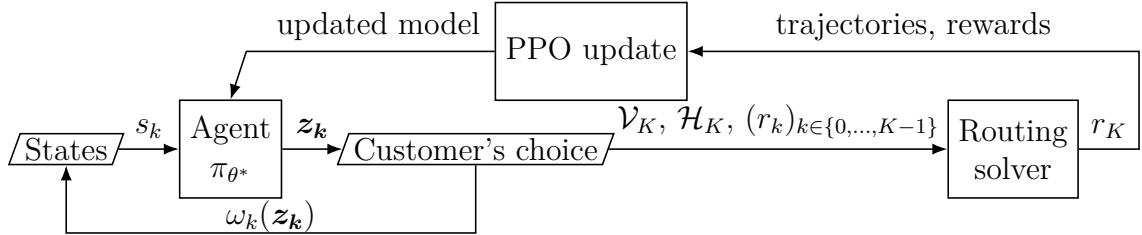
\begin{figure}[h!]
\centering
\begin{tikzpicture}[
 node distance = 5mm and 7mm,every text node part/.style={align=center},
start chain = going right,
lab/.style={inner sep=0pt},
mdl/.style = {shape=ellipse, draw,align=center, inner sep=0.6mm,font=\linespread{1}\selectfont, xscale=0.9,yscale=1},
io/.style = {draw, align=center, inner sep=0.6mm,shape = trapezium,trapezium left angle=70, trapezium right angle=110, font=\linespread{1}\selectfont},
process/.style = {draw, align=center, font=\linespread{1}\selectfont, minimum height=1.3cm},
every path/.style = {draw, -Stealth, semithick},
block/.style={rectangle split, draw, rectangle split parts=2,
text width=14em, text centered, rounded corners, minimum height=4em},
container/.style={draw=white, rectangle,inner sep=0.05cm, rounded
corners,fill=yellow!20,minimum height=4.5cm}
]
%%%%%%%%%%%%%%%%   nodes   %%%%%%%%%%%%%%%%%
\node (state) [io] {States};
\node (gat_node) [process, right = 0.8 cm of state] {Agent\\$\pi_{\theta^*}$};
\node (customer_choice) [io, right = 0.8cm of gat_node] {Customer's choice};
\node (tsp_node) [process, right = 4.5cm of customer_choice] {Routing\\solver};

\draw[-Latex]   (state) -- node[midway, above]{$s_k$}  (gat_node); % <-- arrow
\draw[-Latex]   (gat_node) -- node[midway, above]{$\vec{z_k}$}  (customer_choice); % <-- arrow
\draw[-Latex] (customer_choice) -- node[midway, above]{$\mathcal{V}_K$, $\mathcal{H}_K$, $(r_k)_{k\in \{0,\dots, K-1\}}$} (tsp_node); % <-- arrow
%%%%%%%%%%%%%%% Loop arrow %%%%%%%%%%%%%%%%%%%%%%%%%

\coordinate (up) at ($(customer_choice.south)+(0,-9mm)$);
\coordinate (left) at ($(state.south)+(0,-9mm)$);
\draw[-Latex] (customer_choice.south) -- (up) -- node[lab, above] {$\omega_k(\vec{z_k})$} (left) -- (state.south);

\coordinate (bullet3) at ($(tsp_node.east)+(8mm, 0mm)$);
\coordinate (up2) at ($(bullet3.north)+(0,+13mm)$);
\node (ppo_node) [process, left = 6 cm of up2] {PPO update};

\coordinate (left2) at ($(ppo_node.west)+(-3.3cm,0)$);

\draw[-Latex](tsp_node) -- node[midway, above]{$r_K$} (bullet3.north) -- (up2) -- node[midway, above]{trajectories, rewards}(ppo_node.east);

\draw[-Latex] (ppo_node.west) -- (left2) -- node[pos=0, above, xshift=50pt]{updated model} (gat_node.north);

\end{tikzpicture}
\caption{Main components and information flow in the DPO architecture}\label{fig:architecture}
\end{figure}

A central challenge is estimating the downstream impact of pickup point offers on the final routing cost. This difficulty arises for several reasons. First, future customer locations are unknown at the time of decision-making, and customers that appear unfavorable for detours early on may later form clusters that are efficient for routing. Second, before observing all customer locations, it is difficult to identify effective consolidation points. Third, decisions are interdependent across customers: once a pickup point is offered, it may be preferable to reuse the same location for subsequent customers rather than introduce additional stops, even if alternative pickup points are geographically closer. Finally, the total routing cost depends on stochastic customer choices between home delivery and pickup points and can only be evaluated after the order capture period has ended.

To address these challenges, we adopt a DRL approach, which is well-suited for sequential decision-making under uncertainty and delayed cost realization. In reinforcement learning, an agent interacts with an environment by selecting actions based on observed states and receives feedback in the form of costs or rewards. The agent seeks to learn a policy that minimizes the expected cumulative cost over time \citep{sutton2018reinforcement}. 

A key modeling challenge lies in representing the state. The state comprises the spatial relationships among a dynamically changing set of customers and pickup points, resulting in a large, variable information space. To effectively capture relational and geographical structure, we encode the state as a graph with an attention mechanism \citep{bahdanau2014neural}. In graph attention networks (GATs, \citep{velivckovic2017graph}), attention mechanisms dynamically weight information from neighboring nodes. This allows the model to focus on the most relevant entities, such as nearby customers or frequently used pickup points, while still incorporating global information. Rather than manually specifying which nodes are important, the attention mechanism learns task-specific relevance weights directly from data, reducing the risk of discarding critical information or overwhelming the model with irrelevant details.

The agent processes the graph-based state through the GAT encoder to produce low-dimensional node embeddings. For each arriving customer, these embeddings are used to select a pickup point offer. This process is repeated sequentially throughout the order capture phase. Once all customer arrivals and choices are realized, an off-the-shelf routing solver constructs the final delivery route. The resulting routing cost, combined with customer travel costs, defines the episode-level objective. The collected trajectories are then used to update the policy via Proximal Policy Optimization \citep{schulman2017proximal}. Figure~\ref{fig:architecture} summarizes the interaction between the agent, the environment, and the routing solver.

Overall, the proposed algorithmic design combines three key strengths. First, the DRL framework allows tackling the problem's uncertainty, delayed cost realization, and interdependent decisions. Second, the graph-based state representation with attention enables scalable modeling of geographical relationships in variable-size problems while mitigating oversmoothing by selectively propagating relevant information. Third, integrating GATs with PPO yields a stable, data-efficient learning procedure that produces high-quality policies without relying on problem-specific heuristics.

\subsection{Network architecture}\label{sec:network_architecture}

We implement an actor-critic architecture with two separate neural networks: a policy network $\pi_{\theta}(\vec{z}_k|s_k)$ parameterized by $\theta$, and a value network  $V_{\psi}(s_k)$ parameterized by $\psi$. Both networks take as input the graph representation of state $s_k$. The policy network outputs a probability distribution over decisions $\vec{z_k}$ for a given state $s_k$.  The value network outputs the expected cumulative reward for the current state $s_k$.  In the following, we first describe the encoder structure used by both networks, and then describe the decoders specific to each network. Although the graph operations are described for a single state $s = (\mathcal{V}, \mathcal{H}, \mathcal{A})$, they are implemented in batches for efficiency. 

To generate node embeddings, the encoder employs two attentional layers, following the architecture from \citet{velivckovic2017graph}.  Each feature vector $\vec{h_i} \in  \mathcal{H}, i\in \mathcal{V}$ is first passed through a shared linear layer with weight matrix  $W:\mathbb{R}^{|\vec{h_i}|}\rightarrow \mathbb{R}^B$, where $B$ is the size of the embedding, to produce the initial representation $W\vec{h_i}$. For each node $i$, a new representation is computed by aggregating information from its neighbors, weighted by attention coefficients $\alpha_{ij} \in [0,1]$, which indicate the importance of neighbor $j$'s features to the node $i$. Let $\mathcal{V}(i)$ denote the first-order neighbors of a node $i\in\mathcal{V}$; in our problem, this set includes all nodes that must be visited during the fulfillment phase. The representation is computed as:
 \begin{equation}
     \vec{h_i}' = ELU \big(\sum_{j\in\mathcal{V}(i)} \alpha_{ij} W\vec{h_{j}}\big),
 \end{equation}
 where $ELU$ is an exponential linear unit activation function. The attention coefficients $\alpha_{ij}$ are computed for each node pair $(i,j)$ using  a learnable parameter vector $\vec{a}:\mathbb{R}^{2B} \rightarrow \mathbb{R}$, and are normalized across all neighbors of the node $ i\in \mathcal{V}$:
 \begin{equation}
     \alpha_{ij} =\frac{\exp\left(\text{LeakyReLU}\left(\vec{a}^T[ W\vec{h}_i \|W\vec{h}_j]\right)\right)}{\sum_{k\in\mathcal{V}(i)}\exp\left(\text{LeakyReLU}\left(\vec{a}^T[W\vec{h}_i \| W\vec{h}_k]\right)\right)}.
 \end{equation}
 where $\|$ denotes the concatenation,  and LeakyReLU is the leaky rectified linear unit activation function.  To stabilize learning, a multi-head attention approach is used, where the features from $R=4$ independent replicas of attentional mechanisms are concatenated:
 \begin{equation}
     \vec{h_i^{(1)}} =  \|_{\rho=1}^R ELU \big(\sum_{j\in\mathcal{V}(i)} \alpha_{ij}^{\rho} W^{\rho} \vec{h_{j}}\big), 
 \end{equation}
 where  $\alpha_{ij}^{\rho}$ and  $W^{\rho}$ are the attention coefficients and the weight matrix for the $\rho$-th replica, respectively.  
 
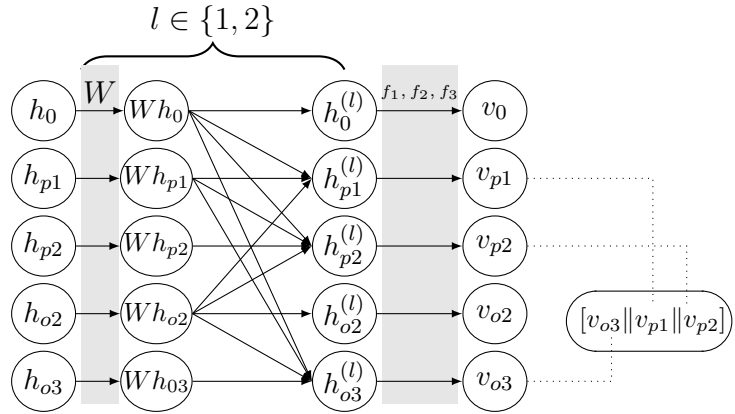
\begin{figure}[h]
    \centering
       \begin{tikzpicture}[
state_node/.style={ellipse, draw=black, minimum height=0.8cm, minimum width=0.85cm, inner sep=-3pt, font=\small}]

%state_node/.style={circle,  minimum size=0.65cm, inner sep=0pt, draw=black}]%
%%%%%%%%%%%%%%%%%%%%%%%%  fulfillment phase %%%%%%%%%%%5
 \begin{scope}[scale=1]
    \node[state_node] (d) at  (11.5, 3.6) {$h_0$};
    \node[state_node] (p1) at (11.5, 2.7) {$h_{p1}$};
    \node[state_node] (p2) at (11.5, 1.8) {$h_{p2}$};
    \node[state_node] (o2) at (11.5, 0.9)  {$h_{o2}$};
    \node[state_node] (o3) at (11.5, 0) {$h_{o3}$};
    %%%%%%%%% 

    \node[state_node] (gd) at  (13, 3.6) {\footnotesize{$Wh_0$}};
    \node[state_node] (gp1) at (13, 2.7) {\footnotesize{$Wh_{p1}$}};
    \node[state_node] (gp2) at (13, 1.8) {\footnotesize{$Wh_{p2}$}};
    \node[state_node] (go2) at (13, 0.9)  {\footnotesize{$Wh_{o2}$}};
    \node[state_node] (go3) at (13, 0) {\footnotesize{$Wh_{03}$}};

    \node[state_node] (hld) at  (15.5, 3.6) {$h_0^{(l)}$};
    \node[state_node] (hlp1) at (15.5, 2.7) {$h_{p1}^{(l)}$};
    \node[state_node] (hlp2) at (15.5, 1.8) {$h_{p2}^{(l)}$};
    \node[state_node] (hlo2) at (15.5, 0.9)  {$h_{o2}^{(l)}$};
    \node[state_node] (hlo3) at (15.5, 0) {$h_{o3}^{(l)}$};

    \node[state_node] (ld) at  (17.5, 3.6) {$v_0$};
    \node[state_node] (lp1) at (17.5, 2.7) {$v_{p1}$};
    \node[state_node] (lp2) at (17.5, 1.8) {$v_{p2}$};
    \node[state_node] (lo2) at (17.5, 0.9) {$v_{o2}$};
    \node[state_node] (lo3) at (17.5, 0) {$v_{o3}$};
 
    %%%%%%%%%%% on top and gray areas %%%%%%%%%%%%
    \draw[decorate,decoration={brace,amplitude=10pt,raise=4pt},thick] (gd.north)++(-1,0)  -- (hld.north) node[midway, above=12pt] {$l\in\{1,2\}$};

    \fill[gray!20] (12, -0.3) rectangle (12.5, 4.2);
    \node at (12.25,3.85) {$W$};

    \fill[gray!20] (16, -0.3) rectangle (17, 4.2);
    \node at (16.5,3.85) {\tiny{$f_1, f_2, f_3$}};
        
   % Define auxiliary nodes
    \node[inner sep=-5pt] (A) at (20.3,0.4) {};
    \node[inner sep=-5pt] (C) at (18.8,1.2) {};
    
    % Define nodes B and D relative to A and C, respectively
    \node[inner sep=-5pt] (B) at ($ (A) + (0,0.8) $) {}; % B is directly above A
    \node[inner sep=-5pt] (D) at ($ (C) + (0,-0.8) $) {}; % D is directly below C
    
    % Draw arcs and lines using the auxiliary nodes
    
    \draw [-] (D) -- (A) ++(0.4,0);
    \draw [-] (C) -- (B) ++(0.4,0); % Ensure B is defined relative to C
    \draw (A)++(-0.05,0) arc (-90:90:0.4);
    \draw (C)++(+0.05,0) arc (90:270:0.4);
    \node[] at (19.6, 0.8)  (hhhbar) {\footnotesize{$\left[v_{o3}\|v_{p1}\|v_{p2}\right]$}};

    \draw [dotted] (lp1) -| (19.6, 1);
    \draw [dotted] (lp2) -| (20.05, 1);
    \draw [dotted] (lo3) -| (19.05, 0.6);
    
    %%%%%%%%% 1-2 connections %%%%%%%%%%%%%
    
    \draw[->,  >=latex] (d) -- (gd);
    \draw[->,  >=latex] (p1) -- (gp1);
    \draw[->,  >=latex] (p2) -- (gp2);
    \draw[->,  >=latex] (o2) -- (go2);
    \draw[->,  >=latex] (o3) -- (go3);

    %%%%%%%%%  2-3 connections %%%%%%%%%%%%%

    \draw[->,  >=latex] (gd.east) -- (hld.west);
    \draw[->,  >=latex] (gd.east) -- (hlp1.west);
    \draw[->,  >=latex] (gd.east) -- (hlp2.west);
    \draw[->,  >=latex] (gd.east) -- (hlo3.west);
    
    \draw[->,  >=latex] (gp1.east) -- (hlp1.west);
    \draw[->,  >=latex] (gp1.east) -- (hlp2.west);
    \draw[->,  >=latex] (gp1.east) -- (hlo3.west);
    
    \draw[->,  >=latex] (gp2.east) -- (hlp2.west);
    
    \draw[->,  >=latex] (go2.east) -- (hlo2.west);
    \draw[->,  >=latex] (go2.east) -- (hlp1.west);
    \draw[->,  >=latex] (go2.east) -- (hlp2.west);
    \draw[->,  >=latex] (go2.east) -- (hlo3.west);
    
    \draw[->,  >=latex] (go3.east) -- (hlo3.west);

    %%%%%%%%%  3-4 connections %%%%%%%%%%%%%
    \draw[->,  >=latex] (hld) -- (ld);
    \draw[->,  >=latex] (hlp1) -- (lp1);
    \draw[->,  >=latex] (hlp2) -- (lp2);
    \draw[->,  >=latex] (hlo2) -- (lo2);
    \draw[->,  >=latex] (hlo3) -- (lo3);
    
    % \draw[->,  >=latex] (lp1.east) -- (ggg.west);
    % \draw[->,  >=latex] (lp2.east) -- (ggg.west);
    % \draw[->,  >=latex] (lo3.east) -- (ggg.west);
\end{scope}

\end{tikzpicture}
   \caption{Structure of encoder  for input of state example from Figure~\ref{fig:state_example}}\label{fig:layers}
\end{figure}

The second attentional layer, with an architecture similar to that of the first, takes the embeddings $\vec{h^{(1)}}_i$ from the first layer as input and produces the final embeddings $\vec{h^{(2)}}_i$.  These embeddings are then passed through three shared fully connected layers: $f_1:\mathbb{R}^{B \times R} \rightarrow \mathbb{R}^{H}$, $f_2:\mathbb{R}^{H} \rightarrow \mathbb{R}^{H}$, and $f_3:\mathbb{R}^{H} \rightarrow \mathbb{R}$, where $H$ is the parameter of size of hidden layers. The first two layers use a hyperbolic tangent activation. The resulting values $v_i$  of nodes representing pickup points and the new customer are concatenated into a vector  $\|_{i\in\mathcal{M}\cup \{o_k\}}\vec{v_i}$. A schematic illustration of the encoder for both networks is shown in Figure~\ref{fig:layers}.

Finally, for the policy network, the concatenated vector $\|_{i\in\mathcal{M}\cup \{o_k\}}\vec{v_i}$ is passed through a softmax function to produce valid action probabilities.  For the value network,  the concatenated vector passed through a fully connected layer $f_{v2}: \mathbb{R}^{|\mathcal{M} + 1|}\rightarrow  \mathbb{R}$. 

\subsection{Proximal policy optimization}\label{sec:ppo}

To train the policy and value networks, we use the proximal policy optimization (PPO) algorithm, a model-free, on-policy reinforcement learning method. This section provides an overview of the algorithm, following the work of \citet{schulman2017proximal}. In reinforcement learning, it is conventional to maximize rewards rather than minimize costs. Therefore, in this section, rewards are defined as the negation of the cost-to-go functions described in Section~\ref{sec:cost_to_go}, and the objective is defined as the negation of objective \eqref{eq:objective}.

Training proceeds episodically, with the agent collecting multiple trajectories. Each trajectory consists of a sequence of $N^{\text{steps}}$ episodes representing arrivals of consecutive new orders. During an episode, the agent observes the current state $s_k$ and samples an action $\vec{z_k}$ based on the distribution obtained by policy $\pi_{\theta}(\vec{z_k}|s_k)$, determining which pickup point to offer for the new order. The environment then returns the associated cost-to-go, $r_k(s_k, \vec{z_k}, \omega_k(\vec{z_k}))$, based on exogenous information about the customer's choice and the details of the next order. The state then transitions to $s_{k+1}$.  An experience tuple $(s_k, \vec{z_k}, r_k(s_k, \vec{z_k}, \omega_k(\vec{z_k})), done_k)$ is stored to be used for network updates, where $done_k$ indicates the termination of the capture phase.

Once trajectories are collected, batches of $N^{\text{batch}}$ episodes are sampled from the dataset and used to update the policy and value functions via stochastic gradient ascent with the Adam optimizer \citep{kingma2014adam}. The specificity of PPO is that it constrains the magnitude of policy update to improve learning stability. With this objective, the policy update is performed by minimizing a surrogate objective that combines three terms. The first is the clipped cost ($L_k^{CLIP}(\theta)$), which prevents the updated policy $\pi_{\theta}$ from diverging excessively from the previous policy $\pi_{\theta^{\text{old}}}$. This term is defined as:
\begin{equation}
    L^{CLIP}(\theta) = \mathbb{E}_k\left[\min \left(\frac{\pi_{\theta}(z_k|s_k)}{\pi_{\theta^{\text{old}}}(z_k|s_k)} \hat{A}_k, 
    \text{clip}\left(\frac{\pi_{\theta}(z_k|s_k)}{\pi_{\theta^{\text{old}}}(z_k|s_k)}, 1-\epsilon^{\text{clip}}, 1+\epsilon^{\text{clip}}\right)\hat{A}_k \right)\right],
\end{equation}
where the clip function sets to zero the gradient whenever the probability ratio $ \frac{\pi_{\theta}(z_k|s_k)}{\pi_{\theta^{\text{old}}}(z_k|s_k)}$, is outside of the interval $[1-\epsilon^{\text{clip}}, 1+\epsilon^{\text{clip}}]$. The expectation is taken over the minimum of the clipped and unclipped advantages to obtain a lower bound on the loss function. The advantage function $\hat{A}$ is computed using the generalized advantage estimation  $GAE(\lambda)$ \citep{schulman2015high} with discount factors $\gamma$.

The second term is the value function loss $L_k^{VF}(\theta)$, a squared-error term that penalizes the difference between the predicted value $V_{\psi}(s_k)$ and the actual return $V_k^{\text{target}}$. The third term, $S[\pi_{\theta}(s_k)]$, encourages exploration during training \citep{williams1992simple, mnih2016asynchronous}. The overall objective function is given by:
\begin{equation}
    L_k(\theta) = \mathbb{E}\left[L_k^{CLIP}(\theta) + c_1 L_k^{VF}(\theta) + c_2 S[\pi_{\theta}(s_k)]\right],
\end{equation}
where $c_1$ and $c_2$ are weighting coefficients.

After each update, the trajectories are discarded due to PPO's on-policy nature, and the updated policy then performs new actions.  The algorithm is trained for $N^{\text{total}}$ steps, yielding the final network weights, denoted $\theta^*$. For any given state $s_k$, the learned policy selects the decision $\vec{z_k}$ corresponding to the highest probability. We refer to this policy as $DPO$.

%%%%%%%%%%%%%%%%%%%%%%%%%%%%%%%%%%%%%%%%%%%%%%%%%%%%%%%%%%%%%%%%%%%%%%%
%                           Experimental setup                        %
%%%%%%%%%%%%%%%%%%%%%%%%%%%%%%%%%%%%%%%%%%%%%%%%%%%%%%%%%%%%%%%%%%%%%%%

\section{Experimental setup}\label{section:instance_design_setup}

The section outlines experimental setup for analyzing the proposed approach, including instance design (Section~\ref{section:instances}) and training and testing design (Section~\ref{section:train_test_design}). 

All algorithms and the simulation environment for the experiments were implemented in Python 3.10. The PyTorch Geometric library \citep{Fey2019} was used to implement graph attentional layers, and the Gurobi solver \citep{gurobi} was employed to solve the routing subproblem. The experiments were conducted on a single CPU thread of an AMD EPYC Rome Processor with 64 GB of memory.

\subsection{Instance design\label{section:instances}}

Synthetic instances are generated following a modified approach based on \cite{dellaert2019branch} to simulate an urban service region with varying spatial distribution. To identify settings in which differentiated pickup point offerings are most or least effective, the experimental design varies both spatial and customer behavior parameters.

\subsubsection{Spacial parameters}
\begin{figure}[ht!]
\centering
 \includegraphics[width=0.6\textwidth]{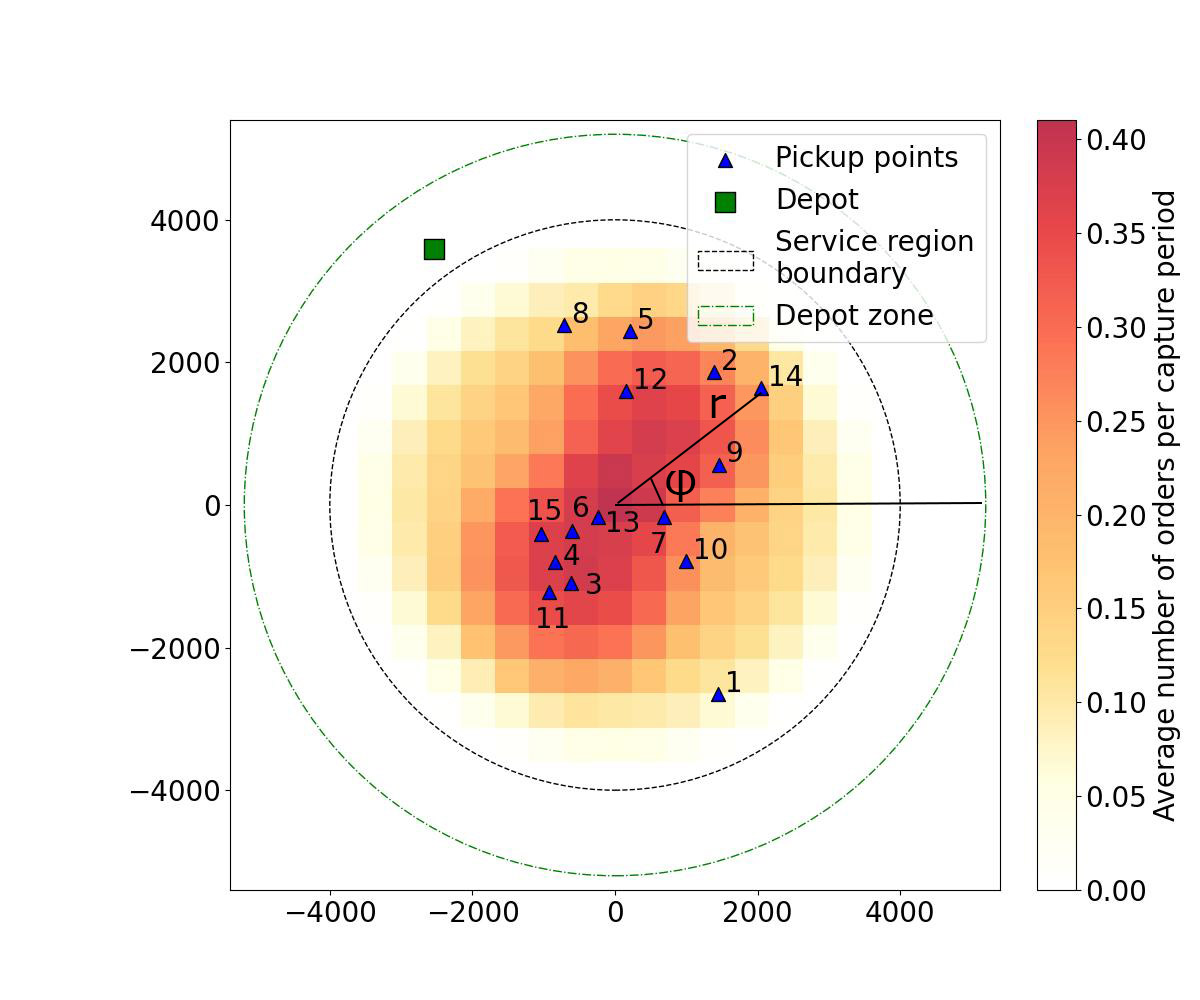}
  \caption{Representation of the delivery service region for $L = 4$ km and $|\mathcal{M}| = 15$ pickup points. The color scheme indicates the average number of orders per order capture phase arriving from the corresponding square area of size $0.5 \text{km} \times 0.5 \text{km}$.}\label{figure:instance_map}
\end{figure}

The locations of customers and pickup points are distributed so that 40\% are within a central circular zone of radius $L$. The remaining 60\% are distributed equally between two smaller but densely populated zones, each with a radius of $0.5L$. These smaller zones have centers randomly positioned between $0.3L$ and $0.8L$ from the service region's center. All three zones may intersect, resulting in a combined distribution of locations. Pickup points follow the same spatial distribution as the customer locations. The depot is located outside the city boundaries, at a distance of $L$ to $1.3L$ from the center of the service region. To determine the locations of vertices, we randomly select a radius $r$ within the specified zone and an angle $\phi$ between $0$ and $2\pi$. Figure~\ref{figure:instance_map} illustrates an example of the resulting spatial distribution of order locations, pickup points, and the depot.

To evaluate how DPO performance varies with pickup point network density, we generate instances with pickup point counts $|\mathcal{M}| = 4, 7, 15, 30$. For each $|\mathcal{M}|$,  five distinct geographical instances are generated,  with random positions of the pickup points and depot. To examine the effect of service region size, all coordinates are scaled for $L\in\{0.5, 1, 2, 4, 8\}$ km. Smaller $L$ represents dense urban regions with short walking distances, while larger $L$ represents more rural regions. 

The resultant 100 spatial instances will be made available at
\href{https://tomvanwoensel.com/instances-used/}{\url{https://tomvanwoensel.com/instances-used/}}.

\subsubsection{Customer behavior parameters}

Order arrivals follow a homogeneous Poisson process during an 8-hour order capture phase, divided into 15-minute periods, with an arrival rate of one order per 15-minute period. This corresponds to approximately 32 orders per day. 

We vary the customer choice parameters for pickup point and home delivery (Equation \eqref{eq:probability_pup}) to examine how different levels of acceptance of pickup point delivery affect DPO performance. This model accounts for both distance-related and non-distance-related factors affecting customer choice.  To capture different behavioral patterns, we define three probability settings: Base, Low, and High. In the Base case, we assume that customers opt for a pickup point over home delivery with a 0.7 probability when the pickup point is 200 meters from their home and with a 0.3 probability when the distance is 4000 meters. In High- and Low-probability settings, these values shift to 0.9 and 0.5 (high) and 0.5 and 0.1 (low), respectively. Table \ref{tab:probability_parameters} presents the corresponding utility parameters for each setting.
\begin{table}[h]
    \centering
\begin{tabular}{llllllllllllllllll}
\toprule
Probability settings & $p_{om}$ ($c_{om} = 200$ m)&  $p_{om}$ ($c_{om} = 4000$ m) & $\beta^{\text{pup}}$& $u_0^{\text{pup}}$& $u_0^{\text{home}}$\\
\midrule
Low & 0.5 & 0.1 & 0.57 & -1.88 & -2.00 \\
Base & 0.7 & 0.3 & 0.45 & -1.06 & -2.00 \\
High & 0.9 & 0.5 & 0.58 &  0.31 & -2.00 \\
\bottomrule
\end{tabular}
 \caption{Utility parameters for customer choice between pickup point and home delivery} \label{tab:probability_parameters}
\end{table}

The customer's choice of transportation modes for order collection (car trip, chained car trip, cycling or walking) is modeled based on the utilities in Equation \eqref{eq:prob_car_np}, with parameter values from \citet{niemeijer2023greener}: $u_0^{\text{car}} = -3.532$, $u_0^{\text{chained}} = -2.944$, $u_0^{\text{cycle}} = -1.593$, $u_0^{\text{walk}} = 0$, $\beta^{\text{car}} = 2.481$, $\beta^{\text{chained}} = 2.398$, $\beta^{\text{cycle}} = 1.845$, $\beta^{\text{walk}} = 0$. Carbon dioxide emission per unit distance is based on \citet{niemeijer2023greener}, with values of $ e^{\text{truck}} = 196$ g of carbon dioxide per kilometer and $ e^{\text{car}} = 116$ g of carbon dioxide per kilometer. 

% \begin{figure}[h]
% \centering
% \begin{tikzpicture}[scale=0.50, transform shape]
%   \draw[white] (0,0) circle (1cm) node[below=1cm] {\textcolor{black}{Area 1}};
%   \draw[thick] (0,0) circle (2cm) node[below=2cm] {Area 2};
%   \draw[thick] (0,0) circle (3cm) node[below=3cm] {Area 3};
%   \draw[thick] (0,0) circle (4cm) node[below=4cm] {};
%   \draw[->] (-5,0) -- (5,0) node[right] {$y$};
%   \draw[->] (0,-5) -- (0,5) node[above] {$x$};
%   \filldraw[fill=gray!30] (0,0) -- (3cm,0) arc (0:30:3cm) -- cycle;
%   \draw (1.5cm,0.5cm) node {$\phi$};
%   \draw (2cm,0) arc (0:30:2cm);
%    \draw[dashed] (30:3cm) -- (30:3cm |- 0,0) node[below right] {$r$};
% \end{tikzpicture}
%  \caption{Schematic representation of the delivery service zone}\label{figure:instances}
% \end{figure}

\subsection{Training and testing design \label{section:train_test_design}}

During training, sequences of new order arrivals are generated randomly, and customer locations are generated according to the spatial distribution described by the spatial instance. To reduce data variability during testing, we simulate 100 identical random sequences of new-order arrivals for all policies, each representing 1 day of delivery planning. For each sequence, the order capture phase is simulated 100 times to account for stochastic customer choices. Note that the simulation over the customer's choice is not done for the perfect information benchmark. The results are then averaged over five geographic instances.

For each problem setting, the \textit{DPO} policy is trained separately. The parametric values used for the PPO algorithm are mostly similar to those of the original work of \citet{schulman2017proximal}, although some adjustments have been made after a grid search. The algorithm is set to train for $N^{\text{total}}=300,000$ steps. Policy updates are performed after a trajectory of $N^{\text{steps}} = 1024$ episodes has been collected. The batches of $N^{\text{batch}}= 128$ episodes are then sampled from the trajectory data for each gradient step with discount factor $\gamma=0.999$, and $GAE(\lambda) = 0.96$. The probability ratio clipping threshold is set to $\epsilon^{\text{clip}} = 0.2$. The weighting coefficients for the surrogate objective terms are $c_1=1$ and $c_2=0.01$. The network parameters are updated using stochastic gradient ascent with a learning rate of $10^{-5}$ and an entropy coefficient of $0.05$. The embedding size is set to $B=64$, and the size of the hidden layers of both the critic and actor is set to $H=128$.  We use $R=4$ independent replicas of the attention mechanism.

Finally, to account for potential sensitivity of neural networks to input scales, we normalize node features: the horizontal and vertical coordinates denoting the location of each node $i$, $d_i = (d_{i}^{\text{horizontal}}, d_{i}^{\text{vertical}})$ are adjusted by adding $L$ and then dividing by the service region's diameter $2L$, and the arrival time $t_i$ by the length of the capture phase $T$. These adjustments scale all values to the range $[0,1]$.

\section{Computational experiments\label{section:computational_experiments}}

This section presents the computational experiments designed to analyze the policy's learned decisions and evaluate the performance of the proposed algorithm. Section~\ref{section:analysis} investigates the behavior of the learned policy under varying operational and customer behavior settings. Section~\ref{section:algorithm_evaluation} compares the algorithmic design with different model architectures. 

\subsection{Analysis of learned decisions\label{section:analysis}}

This section analyzes the decisions made by the differentiated pickup point offering (\textit{DPO}) policy. Our objective is to characterize the operational settings in which the dynamic differentiated pickup point offering is particularly valuable.  We also assess when simpler policies are sufficient to achieve comparable performance. We examine how pickup point offering decisions respond to four factors: \textit{(i)} the number of available pickup points, \textit{(ii)} the size of the service region, \textit{(iii)} customer choice probabilities, and \textit{(iv)} the timing of the decision. To quantify these effects, we compare \textit{DPO} policy with five alternative policies:

\begin{itemize}
   \item \textit{HOME}: a baseline policy in which all customers receive home delivery, used to assess how pickup points can reduce the emissions of deliveries. 
    \item \textit{PERFECT INFORMATION}: a lower-bound benchmark. The benchmark is calculated by solving a deterministic problem under assumptions that  \textit{(i)} the provider knows locations of all customers in advance and \textit{(ii)} the customers certainly choose the pickup point if it is offered, otherwise the home delivery is done.  While the first assumption is partially addressable by \textit{DPO}'s expectation of future states, the second assumption is difficult to enforce in practice. One simulated instance could be solved within 0.5--2000 seconds to a 0.1\% optimality gap.
    \item \textit{UNRESTRICTED}: represents current industry practice, where customers choose between home delivery and any existing pickup point based on their preference (distance-based utility).
    \item \textit{NEAREST}: each customer is offered their nearest pickup point, reflecting a common modeling assumption in the pickup point literature (e.g., \citep{deutsch2018parcel, janjevic2019integrating, kedia2020locating}).
    \item \textit{DYNAMIC NEAREST}: a simple dynamic heuristic that offers the nearest pickup point during the first 30\% of the order capture phase and thereafter restricts choices to the subset of points selected in that initial period, thereby limiting route fragmentation. The policy is motivated by our experimental insights.
\end{itemize}

The remainder of this section is organized as follows. Sections~\ref{sec:impact_num_pups} and \ref{section:impact_city_radius} examine the effect of the number of available pickup points and the size of the service region on the performance of the \textit{DPO} policy compared to alternative policies.  Section~\ref{sec:impact_choice_prob} examines the impact of the probability of customer choice on the delivery emissions of the \textit{DPO} policy. Section~\ref{sec:impact_time_period} analyzes how the \textit{DPO} policy's decisions evolve throughout the order capture phase.

\subsubsection{Impact of the number of available pickup points\label{sec:impact_num_pups}}
This section compares the performance of the \textit{DPO}  against the alternative policies under varying numbers of available pickup points  $|\mathcal{M}|\in\{4, 7, 15, 30\}$. The service region radius is fixed at $L=4$ km, and all parameters related to customer choices and the order arrival process are set to their base-case values.   Table \ref{tab:nr_pups} reports for each policy and the number of available pickup points $|\mathcal{M}|$ the emissions of trucks and customers, the average number of pickup points visited per route, the share of orders delivered to pickup points, and the average distance customers travel to the offered pickup point.  

Overall, the results show that increasing the number of pickup points improves the potential to reduce delivery emissions. However, realizing this potential requires carefully selecting pickup points to consolidate orders at only a small subset per route. Without such selectivity, some policies can offset the consolidation benefits of pickup point delivery by increasing customer travel or even increasing emissions. This highlights the importance of \textit{DPO} for instances with many pickup points.

\begin{table}[h]
    \centering
\scalebox{0.8}{
\begin{tabular}{llrrrrrr}
\toprule
\multirow[b]{2}{*}{Policy}&
\multirow[b]{2}{*}{ $|\mathcal{M}|$}&
\multicolumn{3}{c}{Carbon dioxide emissions (g)} & 
 \multirow[b]{2}{*}{\shortstack{Av. number of\\visited\\pickup points}}&
\multirow[b]{2}{*}{\shortstack{\% orders at\\pickup point}}   & 
\multirow[b]{2}{*}{\shortstack{ Av. distance to\\ pickup point (m)}}   \\
\cmidrule(lr){3-5}
  &&Total& Truck & Customers&&&\\
\midrule
\textit{HOME}& 0 & 8386 & 8386 & 0 & 0.0 & 0.0 & - \\
\midrule
 & 4 & 5837 & 3888 & 1949 & 2.3 & 51.3 & 1488 \\
 \multirow{1}{*}{\textit{PERFECT}}  & 7 & 5858 & 4456 & 1401 & 2.7 & 48.2 & 1301 \\
 \multirow{1}{*}{\textit{INFORMATION}} & 15 &  5490 & 4082 & 1408 & 4.3 & 54.5 & 1173 \\
 & 30 & 5318 & 3889 & 1429 & 4.1 & 54.1 & 1208 \\
\midrule
\multirow{4}{*}{\textit{DPO}}  & 4 & 7862 & 7023 & 839 & 2.5 & 29.2 & 1406 \\
 & 7 & 7767 & 6978 & 789 & 2.9 & 33.9 & 1255 \\
 & 15 & 7553 & 6809 & 744 & 3.5 & 34.7 & 1192 \\
 & 30 &  7540 & 6772 & 590 & 3.6 & 34.5 & 1252 \\
\midrule
\multirow{4}{*}{\textit{UNRESTRICTED}} & 4 & 9802 & 6590 & 3212 & 3.9 & 45.4 & - \\
 & 7 & 10220 & 6666 & 3555 & 6.3 & 50.9 & - \\
 & 15 & 11006 & 6861 & 4145 & 10.3 & 57.2 & - \\
 & 30 & 11964 & 7652 & 4312 & 13.6 & 60.0 & - \\
\midrule
 \multirow{4}{*}{\textit{NEAREST}}  & 4 & 7853 & 7014 & 839 & 3.6 & 37.9 & 1209 \\
 & 7 & 7879 & 7276 & 602 & 5.4 & 38.6 & 996 \\
 & 15 & 7778 & 7425 & 353 & 8.2 & 41.1 & 697 \\
 & 30 & 8140 & 7937 & 203 & 10.2 & 42.4 & 493 \\
\midrule
& 4 & 7983 & 6892 & 1091 & 2.3 & 35.7 & 1481 \\
\textit{DYNAMIC}  & 7 & 7951 & 7023 & 928 & 2.7 & 35.7 & 1335 \\
\textit{NEAREST}  & 15 & 7719 & 6836 & 883 & 3.2 & 37.1 & 1261 \\
& 30 & 7875 & 7074 & 800 & 3.6 & 37.7 & 1162 \\
\bottomrule
\end{tabular}}
 \caption{Impact of the number of available pickup points $|\mathcal{M}|$ on the performance of \textit{DPO} and alternative decision policies} \label{tab:nr_pups}
\end{table}

In more detail, the \textit{PERFECT INFORMATION} benchmark provides a lower bound on emissions. Compared with home-only delivery, employing pickup points reduces total emissions by approximately 30\% (from 8,386 g to below 5,858 g). This reduction is achieved by nearly halving truck emissions, at the cost of introducing customer emissions, which now account for one-third of total emissions. As $|\mathcal{M}|$ increases, total emissions decrease due to better consolidation opportunities. However, only a small subset of pickup points (2--4) is visited per route, indicating diminishing returns from adding additional pickup locations. 

In more realistic settings, the \textit{DPO} reduces the total emissions by 10\% compared to home-only delivery when $|\mathcal{M}|  = 15$ (from 8,386 g to 7,553 g). Larger $|\mathcal{M}|$ yields more options for consolidating orders and reducing total emissions, both in routing and customer emissions. Yet, the number of pickup points actually visited per route remains low (up to 4), showing that, similarly to  \textit{PERFECT INFORMATION}, the \textit{DPO} selectively exploits spatial flexibility rather than uses all locations.  

The \textit{UNRESTRICTED} policy shows the opposite behavior. As more pickup points are available, customers increasingly choose pickup delivery (from 45.4\% to 60.0\% of orders) at distinct pickup point locations, requesting visits to more than 13 pickup points when $|\mathcal{M}|  = 30$. As a result, the total emissions substantially increase from 9,802 g at $|\mathcal{M}|  = 4$ to 11,964 g at $|\mathcal{M}|  = 30$ and \textit{UNRESTRICTED} to have the highest total emissions across all policies. This arises from the structure of the choice model: when multiple pickup points are offered, each contributes positive utility to pickup delivery relative to home delivery. As a result, the probability that a customer selects the pickup point for delivery increases. 

The \textit{NEAREST} is the best performing policy when $|\mathcal{M}|$ is small, but the emissions increase significantly as the number of pickup points increases. Truck emissions rise from 7,014 g to 7,937 g as $|\mathcal{M}|$  grows, driven by the need to visit many distinct pickup point locations. Although up to 42.4\% of orders are served at pickup points, the need to visit more than 8 pickup point locations at $|\mathcal{M}|\geq 15$ offsets the savings in customer travel. The \textit{DYNAMIC NEAREST} heuristic partially mitigates this effect by restricting later decisions to a subset of the initially selected pickup points, limiting the number of visited locations, and thus reducing routing emissions. Consequently, for large $|\mathcal{M}|$, \textit{DYNAMIC NEAREST} outperforms \textit{NEAREST} in terms of emissions. However, this improvement comes at the cost of substantially longer average customer travel distances to the offered pickup point (1261 m vs. 697 m for $|\mathcal{M}| = 15$).

\begin{figure}[h]
\centering
 \includegraphics[width=0.6\textwidth]{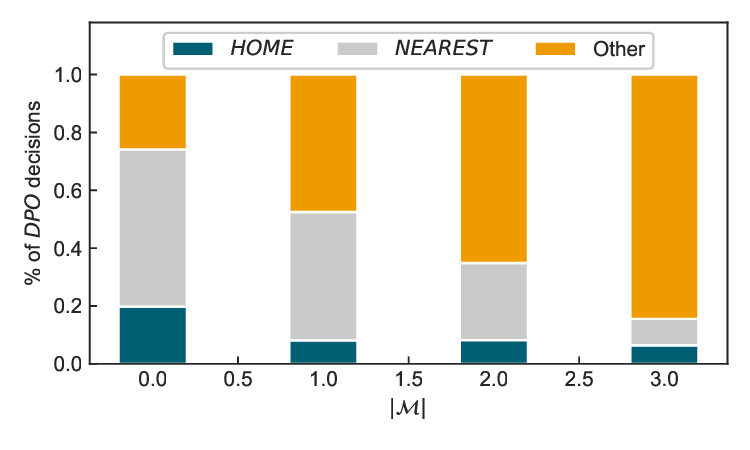}
  \caption{Proportion of \textit{DPO} policy decisions aligning with \textit{HOME} and \textit{NEAREST} policies}\label{figure:nr_pups_aligned_decisions}
\end{figure}

Overall, the table shows that while simpler policies perform reasonably well when few pickup points exist, \textit{DPO} becomes increasingly beneficial as $|\mathcal{M}|$ grows. Figure~\ref{figure:nr_pups_aligned_decisions} illustrates this trend: with only four pickup points, 55\% of \textit{DPO} decisions align with \textit{NEAREST} and 20\% align with \textit{HOME}, whereas those percentages fall below 30\% and 6\%, respectively, when more than 15 pickup points are available.

\subsubsection{Impact of service region's size \label{section:impact_city_radius}}

This section examines the influence of the service region size on the performance of pickup point offering policies, varying the service area size ($L \in \{0.5, 1, 2, 4, 8\}$ km). Because the order-arrival process is the same across all instances, a smaller $L$ corresponds to a more densely populated area. In the experiment, the number of available pickup points is fixed at $|\mathcal{M}| = 15$, and customer choice parameters are set to their base-case values.  We compute the emission reduction achieved by a policy $\pi\in\{ \textit{DPO}, \textit{DYNAMIC NEAREST}, \textit{NEAREST}, \allowbreak\ \textit{UNRESTRICTED}\}$ or by the \textit{PERFECT INFORMATION} benchmark  relative to home-only delivery:
\begin{align*}
 \text{Emission reduction(\%)}= \frac{r(HOME) - r(\pi)}{r(HOME)} \times 100,
\end{align*}
where $r(\cdot)$ denotes total carbon dioxide emissions over the fulfillment process. 

\begin{figure}[h]
\centering
  \subfloat[]{\includegraphics[width=0.32\textwidth]{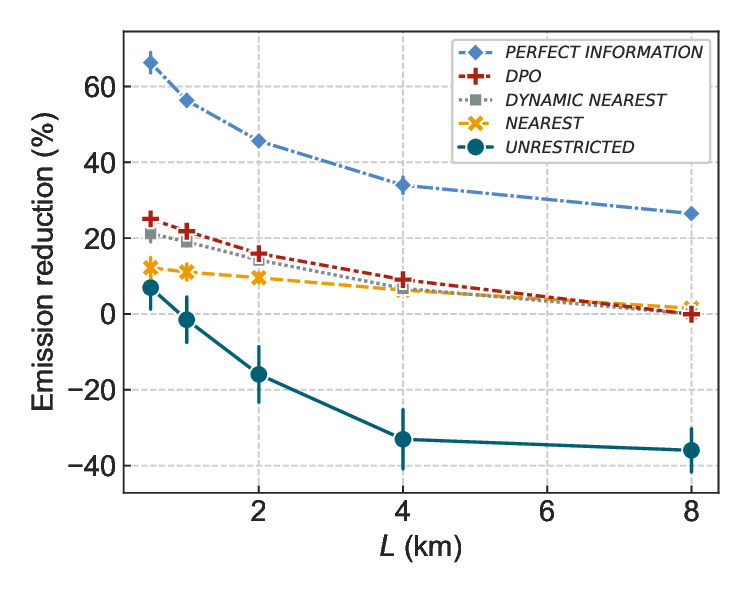}\label{figure:radius_impact_a}}
  \subfloat[]{\includegraphics[width=0.32\textwidth]
  {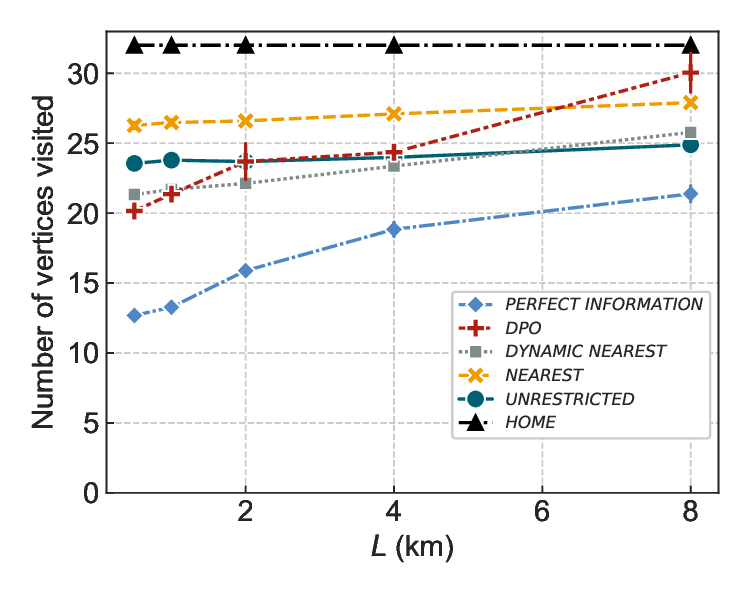}\label{figure:radius_impact_c}}
   \subfloat[]{\includegraphics[width=0.32\textwidth]{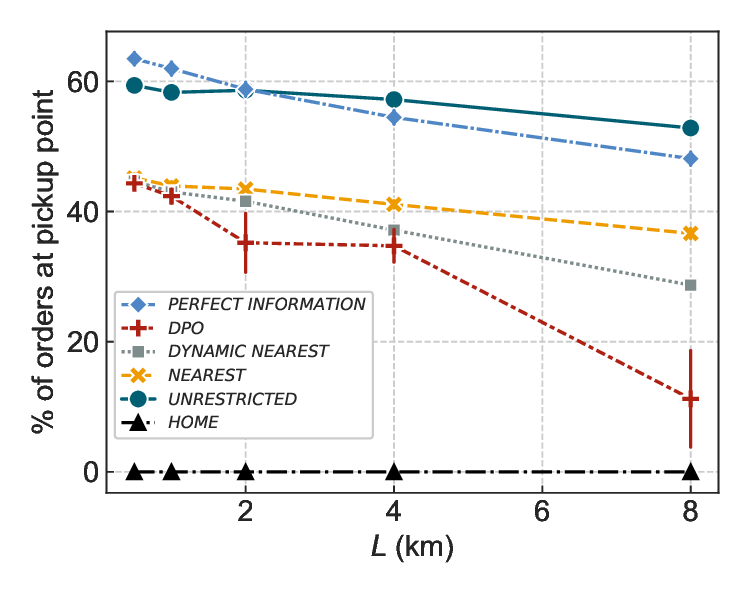}\label{figure:radius_impact_b}}
  \caption{Impact of service region size on the performance of pickup point offering policies. Whiskers indicate one standard deviation across 5 instances, with 10000 simulations per instance.}\label{figure:radius_impact}
\end{figure}

Figure~\ref{figure:radius_impact} shows the impact of service region size $L$ on \textit{(a)} emission reduction relative to home delivery, \textit{(b)} the number of vertices visited per route (including customer homes and pickup points), and  \textit{(c)} the percentage of orders served at pickup points.

The results show that, for all policies, including the \textit{PERFECT INFORMATION} bound, the environmental advantage of pickup point delivery over home delivery declines monotonically as $L$ increases (Figure~\ref{figure:radius_impact_a}). Larger $L$ also leads to fewer orders served via pickup points (Figure~\ref{figure:radius_impact_b}), and to delivery routes that visit more vertices (Figure~\ref{figure:radius_impact_c}). This behavior reflects the impact of distance on customer choice between pickup points and home delivery, as well as on the choice of transportation mode.

In densely populated areas, pickup points are located near customers, making on-demand delivery attractive. As such, when $L$ is less than 1 km, more than 42\% of customers opt for pickup point delivery under all considered policies (Figure~\ref{figure:radius_impact_b}). However, policies differ significantly in their effectiveness in exploiting this opportunity. As shown in Figure~\ref{figure:radius_impact_c}, \textit{HOME} policy requires visiting all customer locations, resulting in an average of 32 visited vertices per route.  The \textit{NEAREST} policy reduces the number of home deliveries but requires visiting many distinct pickup points, resulting in an average of 25 vertices per route. This is because different customers are directed to different locations, limiting consolidation benefits.   In contrast, the \textit{DPO} policy consolidates orders at only 1--2 pickup points,  reducing the number of visited vertices to an average of 20. The strongest consolidation is achieved by the \textit{PERFECT INFORMATION} benchmark, with only 13 vertices per route.

\begin{figure}[h]
\centering
 \includegraphics[width=0.4\textwidth]{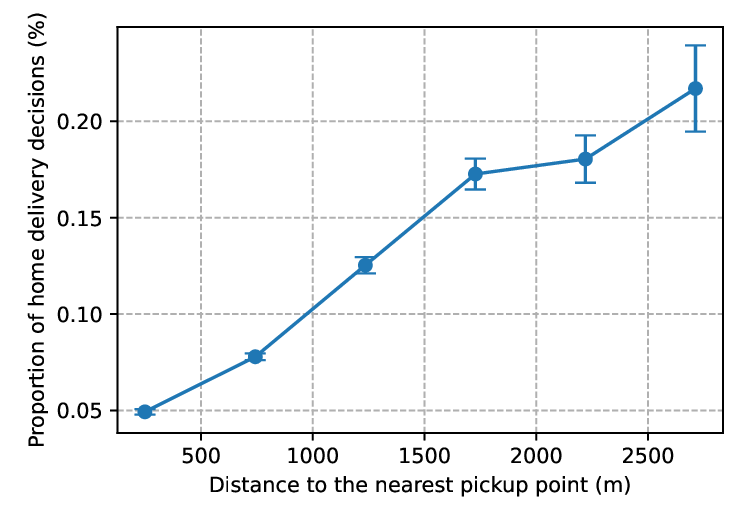}
  \caption{Proportion of customers who were offered home-only delivery. Whiskers indicate one standard deviation across 5 instances, with 10000 simulations per instance.}\label{figure:impact_distance_to_pup_with_error_bars}
\end{figure}

As the service region grows, offering pickup points becomes less environmentally beneficial. Customer travel emissions grow more rapidly with $L$ than truck emissions do, because customers are more likely to opt for round-trip car journeys to reach distant pickup points.  Notably, the \textit{UNRESTRICTED} policy generates emissions that are 38\% higher than home delivery when
 $L = 8$ km.   Figure~\ref{figure:impact_distance_to_pup_with_error_bars} illustrates how \textit{DPO} decisions adapt to avoid inefficient pickup point usage when customer travel emissions outweigh consolidation benefits. For instances with $L=2$ km, the figure shows that as the distance to the nearest pickup point increases, a higher proportion of \textit{DPO} decisions align with the \textit{HOME} policy.

Overall, these observations suggest that pickup point offering policies have a significant potential to reduce total emissions in smaller service regions. In these settings, \textit{DPO} effectively reduces the number of vertices the truck visits, and customers are more likely to use non-polluting transportation for order collection. However, as distances increase, the benefits of pickup point delivery diminish, and offering unrestricted access to pickup points can even lead to higher emissions than home-only delivery. These findings align with the existing literature \citep{mommens2021delivery, niemeijer2023greener}, indicating that pickup points are more sustainable in urban than in rural regions.

\subsubsection{Impact of customer's choice probability\label{sec:impact_choice_prob}}
This section examines how customers' choice of pickup point over home delivery affects the performance of  \textit{DPO}. Table \ref{tab:impact_probability} presents results for high-, base-, and low-probability settings. The service region size is fixed at $L=4$ km, and the number of available pickup points is fixed at $|\mathcal{M}|=15$. 

\begin{table}[h!]
    \centering
    \scalebox{0.9}{
\begin{tabular}{crrrcccc}
\toprule
\multirow[b]{2}{*}{\shortstack{Probability\\settings}}  &\multicolumn{3}{c}{Carbon dioxide emissions (g)}& \multirow[b]{2}{*}{\shortstack{\% orders at\\pickup point}}   & 
\multirow[b]{2}{*}{\shortstack{ Av. distance to\\ pickup point (m)}} \\
\cmidrule{2-4} \cmidrule{7-8}
&Total & Truck& Customer& & \\
\midrule
High & 7498 & 6797 & 701 & 39.6 & 1161  \\
Base & 7553 & 6809 & 744 & 34.7 & 1092  \\
Low & 8142 & 7542 & 600 & 21.2 & 1054  \\
\bottomrule
\end{tabular}}
 \caption{Settings for probability of customer's choice between home and pickup point delivery} \label{tab:impact_probability}
\end{table}

As expected, high-probability settings lead to the best performance of \textit{DPO}. Total emissions decrease from 8,142 g in low-probability settings to 7,498 g in high-probability settings, approaching the \textit{PERFECT INFORMATION } benchmark and substantially improving over the home-only policy (8,386 g; see Table \ref{tab:nr_pups}).  This improvement arises from two factors allowing better order consolidation.  First, the higher likelihood of customers to opt for pickup point delivery results in an increase in the percentage of orders served at pickup points, rising from 21.2 \% in low-probability settings to 39.6\% in high-probability settings. Second, high-probability settings enable the use of pickup points located farther from customers. This is reflected in an increase in the average customer travel distance, which grows from 1054 meters to 1161 meters.

\subsubsection{Change in \textit{DPO} decisions throughout order capture period \label{sec:impact_time_period}}

\begin{figure}[h!]
  \subfloat[]{\includegraphics[width=0.5\textwidth]{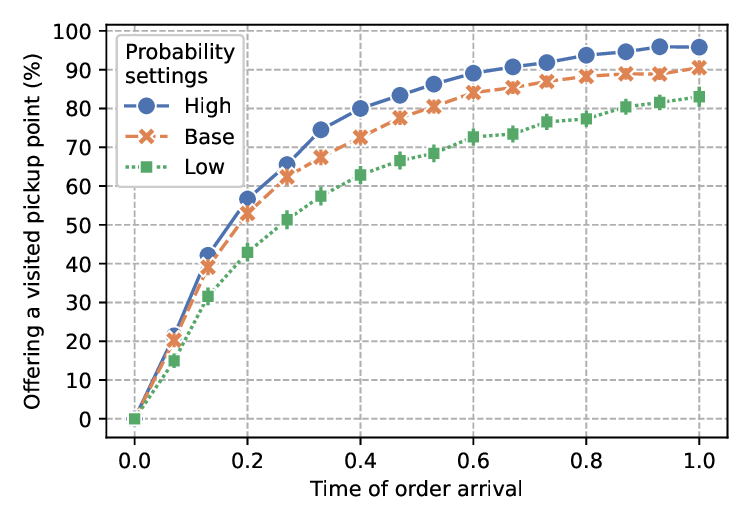}\label{figure:increase_ind_dist_period_a}}
     \subfloat[]{\includegraphics[width=0.5\textwidth]{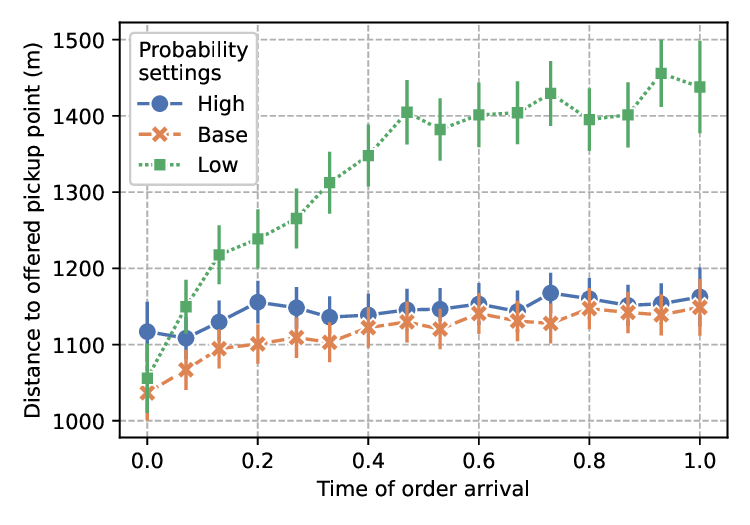}\label{figure:increase_ind_dist_period_b}}
  \caption{Change in the pickup point offering decision throughout the order capture phase. Whiskers indicate one standard deviation across 5 instances, with 10000 simulations per instance.}\label{figure:increase_ind_dist_period}
\end{figure}

This section analyzes the evolution of \textit{DPO} policy decisions throughout the order capture phase. In the experiments, we consider instances with $L=4$ km, $|\mathcal{M}|=15$, and high-, base-, and low-probability settings; the time of order arrival is expressed as a fraction of the capture period $T$. The experiment reports \textit{(i)} the fraction of decisions that offer a pickup point already included in the route for visit (Figure~\ref{figure:increase_ind_dist_period_a}), and \textit{(ii)} the average distance to the offered pickup point (Figure~\ref{figure:increase_ind_dist_period_b}).

Overall, the results indicate that \textit{DPO} decisions exhibit stronger dependence on the system state and greater temporal variability. In contrast, under high-probability settings, decisions are more stable and offer only a small subset of pickup points that perform well on average. This indicates that dynamic decisions and a larger pickup point network are more important when customers are less likely to choose pickup point delivery, and the value of policy adaptivity increases in such settings.

Figure~\ref{figure:increase_ind_dist_period_a} illustrates that under the base probability setting, over 80\% of decisions in the latter half of the order capture phase offer a pickup point already planned to be visited. This proportion rises further in high-probability settings, due to the increased number of visited pickup points. This trend indicates that the set of pickup points to be visited in the fulfillment phase is largely determined early in the order capture phase.  This highlights the anticipatory nature of \textit{DPO} decisions, which prioritize pickup points likely to function as effective consolidation centers for future orders.  Since only a subset of pickup points may function as consolidation centers, the \textit{DPO} policy consistently offers only a subset of all available pickup points. For example, in Figure~\ref{figure:instance_map}, out of 15 available pickup points, only pickup points 4, 9, 11, 12, and 13 were offered in all simulations under high-probability settings. The range of pickup points offered increases as customers become less likely to choose pickup point delivery over home delivery. On average, 5 unique pickup points were offered across all simulations in high-probability settings, 8 in the base case, and 12 in low-probability settings.

Figure~\ref{figure:increase_ind_dist_period_b} illustrates a consistent increase in the distance to the offered pickup point as the order capture phase progresses. This suggests that the policy dynamically adapts to the current state information. For example, it may assign a pickup point that, although located farther from the customer, is already known to be visited during the upcoming fulfillment phase.  This trend is especially evident in low-probability settings. We observe that in the latter half of the order capture phase, the average distance to offered pickup points is 300 meters greater than at the start of the phase.  
This implies that under low-probability conditions, the offering decision is largely influenced by state information, such as prior decisions and customer choices.

\subsection{Algorithmic evaluation\label{section:algorithm_evaluation}}
To assess the contribution of key architectural components of the proposed learning framework, we evaluate the full \textit{DPO} implementation against two variants of the proximal policy optimization (PPO) algorithm. These variants are designed to isolate the effects of the state representation and the attention mechanism:

\begin{itemize}
\item PPO-BASE: a baseline PPO implementation using a flattened state representation (implementation details are provided in Appendix~\ref{appendix:ppo_base});
\item PPO-GNN: a graph-based PPO variant that follows the \textit{DPO} architecture but omits the attention mechanism.
\end{itemize}

\begin{figure}[h!]
\centering
 \includegraphics[width=0.6\textwidth]{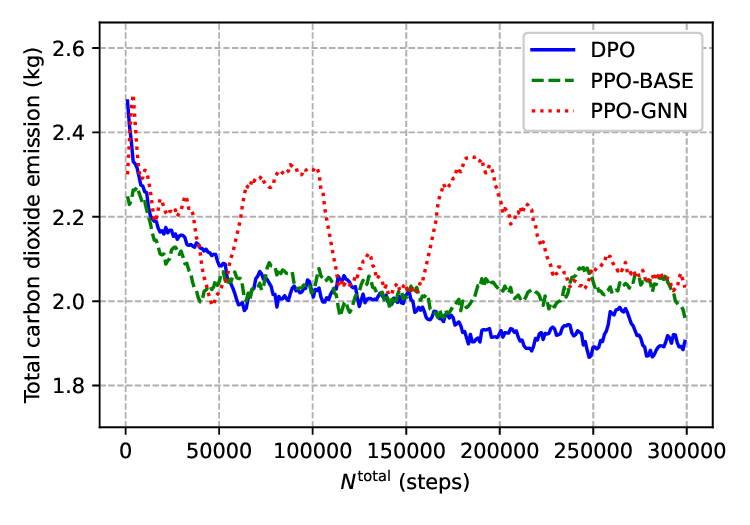}
  \caption{Test objectives throughout learning for different architectures}\label{figure:test_obj_ppo_comparison}
\end{figure}

The results show that the full \textit{DPO} model achieves the lowest objective value and exhibits the most stable learning behavior. PPO-GNN, which excludes the attention mechanism, exhibits noticeably slower convergence and greater training instability. This behavior can be attributed to oversmoothing, resulting in homogenized node embeddings, as observed during training. PPO-BASE converges more stably than PPO-GNN but achieves a higher objective value, suggesting that the flattened state representation lacks the resolution necessary to effectively capture the problem's spatial and relational structure.

Overall, these results indicate that both the graph-based state representation and the attention mechanism contribute to improved learning performance. However, the attention mechanism incurs substantially higher computational cost: a single training step is approximately five times slower than in the PPO-BASE implementation. Consequently, the PPO-BASE model may be preferable in time-constrained settings or for simpler problem instances. In contrast, the attention-based \textit{DPO} architecture is better suited for more complex states.

\subsection{Managerial implications} \label{sec:managerial}
Based on the experimental results presented in this section, we derive several key managerial implications for e-commerce platforms and urban logistics providers, summarized in Table \ref{tab:managerial_implications}.

\begin{table}[h!]
\centering
\caption{Managerial implications for sustainable last-mile delivery}
\label{tab:managerial_implications}
\small
\renewcommand{\arraystretch}{1.3}
\begin{tabular}{p{3.5cm} p{10.5cm}}
\hline
\textbf{Decision area} & \textbf{Operational insight and policy recommendation} \\ \hline
\textbf{Emission perspective} & Platforms must transition from minimizing vehicle-kilometers to a total system emission view. DPO effectively mitigates carbon leakage caused by customer car trips, which often negates the benefits of route consolidation. \\ \hline
\textbf{Choice architecture} & Restricting pickup options to a single strategic recommendation (DPO) does not alienate customers if the recommendation is personalized to the dynamic state of the delivery network; it steers behavior toward lower-emission outcomes. \\ \hline
\textbf{Dynamic value} & The best pickup point is not static. Its value changes based on the spatial density of existing orders. Real-time DRL-based policies are essential to capture these synergies, whereas static nearest-point rules often underperform. \\ \hline
\textbf{Urban planning} & High-density urban centers benefit most from DPO. Planners should focus on a dense network of collection points to ensure the DPO-recommended point remains within the customer's walking utility threshold. \\ \hline
\textbf{Platform strategy} & Differentiated offerings allow platforms to act as environmental intermediaries, using interface design to harmonize individual customer convenience with global emission reduction targets. \\ \hline
\end{tabular}
\end{table}

\section{Conclusions\label{section:conclusion}}
Pickup points are widely recognized as a sustainable alternative to home delivery, as consolidating orders at fewer locations can reduce delivery truck emissions. However, their environmental effectiveness depends critically on customer behavior and how pickup points are offered, since consolidation may offset customer travel emissions. This study examined a differentiated pickup point offering (\textit{DPO}) policy that explicitly accounts for emissions from both delivery operations and customer travel.

We studied this problem as a sequential decision-making setting, in which offering decisions for arriving customers affect future consolidation opportunities and routing outcomes. To address this dynamic interdependence, we adopted a learning-based approach that captures the evolving spatial structure of the system and enables anticipatory decision-making for offerings. Our numerical analysis shows that explicitly accounting for these dynamics is essential for achieving meaningful emission reductions.

The results demonstrate that \textit{DPO} can substantially reduce total emissions relative to commonly used alternatives, including \textit{(i)} home-only delivery, \textit{(ii)} unrestricted pickup point choice, \textit{(iii)} nearest pickup point assignment, and \textit{(iv)} a dynamic heuristic policy. In the base case, \textit{DPO} reduces emissions by up to 9\% compared with home-only delivery and by approximately 2\% on average relative to alternative pickup point offering policies. However, the magnitude of these reductions—and the overall sustainability of pickup point delivery—depends strongly on the delivery context.

In particular, the results indicate that increasing the number of available pickup points does not automatically lead to lower emissions. While a larger pickup point network increases consolidation potential, realizing that potential requires selective, anticipatory offering decisions. Non-selective policies may negate consolidation benefits by increasing customer travel distances and, in some cases, can result in higher emissions than home-only delivery. In contrast, the \textit{DPO} policy consistently concentrates deliveries on a limited subset of pickup points, enabling effective consolidation and more efficient routing. This behavior is especially valuable when customers are less inclined to choose pickup point delivery, as the decision-making process then becomes increasingly dependent on the evolving system state.

The experiments further show that the environmental advantage of pickup point delivery over home delivery declines as the service region expands. Larger service areas reduce customers’ willingness to use pickup points, increase customer travel distances, and lead to delivery routes with more dispersed stops. As a result, pickup points are most effective in compact urban regions, where customers are more likely to rely on non-polluting transport modes and where consolidation opportunities are greatest.

This study highlights the importance of designing pickup point offering policies, rather than merely expanding pickup point networks, to achieve sustainability goals in last-mile delivery. Several directions for future research emerge from this work. One promising extension involves integrating temporal demand management when flexible delivery windows are available, requiring joint decisions on pickup point offerings and delivery timing. Another direction concerns incorporating pickup point capacity constraints and uncertain collection times. Finally, dynamically adjusting pickup point offerings -- analogous to dynamic pricing -- may affect customer perceptions and satisfaction, underscoring the need for further research on behavioral responses to such policies.


\appendix

\section{Implementation details of PPO-BASE}\label{appendix:ppo_base}

PPO-BASE is a standard implementation of proximal policy optimization (PPO) using separate fully connected feedforward neural networks for the actor and the critic, following \citet{schulman2017proximal}. Since this architecture requires a fixed-size input, the original state representation $s_k = (\mathcal{V}_k, \mathcal{H}_k, \mathcal{A}_k)$ is transformed into a flattened state vector of constant dimension, denoted by $s^{F}_k$.

The flattened state consists of three components:
\begin{equation}
    s^F_k = (t_{k}, \text{flat}(G(\{o_k\})), \text{flat}(G(\{i\in \mathcal{V}_k| mv_i = 1\})),
\end{equation}
where $t_{k}$ is the time of the new order's arrival, $\text{flat}(G(\{o_k\}))$ represents the location of the new order, and $\text{flat}(G(\{i\in \mathcal{V}_k| mv_i = 1\})$ represents the locations of nodes that must be visited. %, (c_{o_k m})_{m\in \mathcal{M}}, (mv_m)_{m\in\mathcal{M}})$.

The operator $\text{flat}(G(\mathcal{V}))$ denotes a flattened grid matrix $G$, which encodes the set of vertices $\mathcal{V}$ as a boolean matrix corresponding to a grid of $g \times g$. The element $G_{jj'} = 1$ if there is a vertex $i \in \mathcal{V}$ is located within the grid cell bounded by:

\begin{equation}
    -L + j \frac{2L}{g} \leq x_i < -L + (j+1) \frac{2L}{g}, \quad
    -L + j' \frac{2L}{g} \leq y_i < -L + (j'+1) \frac{2L}{g},
\end{equation}
and $G_{jj'} = 0$ otherwise. 

Both the actor and critic networks use three fully connected layers. The actor network follows the structure:
$\mathbb{R}^{1 + 2 \times g \times g} \rightarrow \mathbb{R}^{H} \rightarrow \mathbb{R}^{H} \rightarrow \mathbb{R}^{|\mathcal{M}| + 1}$,
while the critic network is structured as:
$\mathbb{R}^{1 + 2 \times g \times g} \rightarrow \mathbb{R}^{H} \rightarrow \mathbb{R}^{H} \rightarrow \mathbb{R}$. All layers use ReLU activation functions, except the actor network's output layer, which uses softmax to generate valid probability distributions for action sampling. 

\section*{Declaration of generative AI and AI-assisted technologies in the manuscript preparation process}
During the preparation of this work the authors used ChatGPT (OpenAI) in order to improve readability, grammar and LaTeX formatting. After using this tool/service, the authors reviewed and edited the content as needed and take full responsibility for the content of the published article.

\bibliographystyle{elsarticle-harv} 
\bibliography{lib}
\end{document}